\documentclass[10pt,twocolumn,letterpaper]{article}

\usepackage[pagenumbers]{cvpr}      %

\usepackage{microtype}
\usepackage{graphicx}
\usepackage{booktabs}
\usepackage{multirow}
\usepackage{amsmath}
\usepackage{amssymb}
\usepackage{mathtools}
\usepackage{amsthm}
\usepackage{xspace}

\usepackage{colortbl}
\definecolor{oursbg}{HTML}{FBF3DF}
\definecolor{headbg}{HTML}{EEF1F5}
\definecolor{gainc}{HTML}{A47314}
\definecolor{demph}{gray}{0.5}
\newcommand{\gain}[1]{\textsubscript{\textcolor{gainc}{$\uparrow$#1}}}
\newlength{\gainw}
\newcommand{\gslot}[1]{\makebox[\gainw][l]{#1}}
\newcommand{\oursrow}{\rowcolor{oursbg}}
\newcommand{\headrow}{\rowcolor{headbg}}
\definecolor{cic}{gray}{0.42}
\newcommand{\cint}[2]{\textcolor{cic}{[#1,\,#2]}}
\newcommand{\gap}{read--use gap\xspace}

\newcommand{\ours}{\textsc{Aloud}\xspace}
\newcommand{\obs}[1]{\textbf{Obs.~#1}}
\makeatletter\newcommand{\teasercaption}[1]{\def\@captype{figure}\caption{#1}}\makeatother
\newcommand{\figonecaption}{\textbf{VLMs can read evidence they fail to use, until they say it} (the depicted text is illustrative; the numbers are measured). Left: in one pass at 6\,px, the frozen reader's own attention finds the page it cannot yet read. Middle: at 15\,px, Qwen3-VL-8B recognizes the evidence for 84\% of questions, yet on those questions it answers from the pixels 10.7 points below its own transcript of the same lines: the \gap, which pixels close only at 24\,px with 2.6$\times$ the image tokens (\cref{sec:gap}). Below: thinking readers lose little or nothing, because their thinking recites the evidence. Right: same-budget gain of \ours on long-document QA, and the accuracy behind it at $5\times$, from a context smaller than the compressed pages (\cref{sec:eval}).}

\definecolor{cvprblue}{rgb}{0.21,0.49,0.74}
\usepackage[pagebackref,breaklinks,colorlinks,allcolors=cvprblue]{hyperref}

\def\confName{CVPR}
\def\confYear{2027}

\title{VLMs Read Aloud: The Gap Between Recognizing and Using Visual Text}
\author{Zhiyun Shi\\
Nanyang Technological University\\
{\small ZHIYUN004@e.ntu.edu.sg}}

\begin{document}
\twocolumn[{%
\renewcommand\twocolumn[1][]{#1}%
\maketitle
\begin{center}
\includegraphics[width=\textwidth]{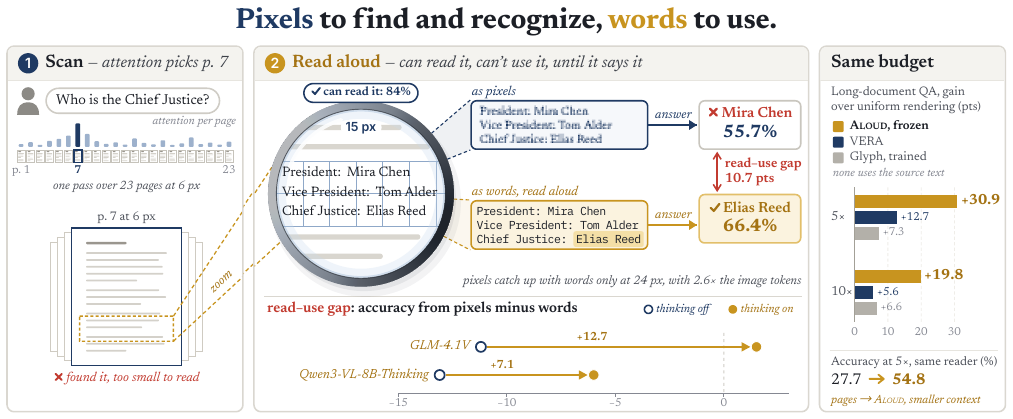}
\vspace{-0.6em}
\teasercaption{\figonecaption}
\label{fig:teaser}
\end{center}
\vspace{0.6em}
}]

\begin{abstract}
Visual-text compression renders long contexts as images and is engineered around how much of the text a vision-language model (VLM) can recover. \textbf{Yet recovered text is not usable text.} VLMs can transcribe the text in an image word for word and still fail to answer from it; given back their own transcript, they answer nearly as well as from the original text. We call this difference the \emph{read--use gap}. Recognizing and using visual text have different resolution requirements: Qwen3-VL-8B's recognition saturates at 15-pixel characters while use catches up only at 24 pixels, and scaling from 2B to 32B raises recognition without narrowing the gap. Putting the evidence into words closes it: thinking VLMs recite the evidence in their thinking, and their own transcript, written without seeing the question, recovers most of what thinking adds. In a preregistered held-out test, the gap ranks how much eight VLMs gain from their own words over pixels (\textbf{Spearman $\boldsymbol{\rho} = \mathbf{0.86}$}), which recognition rates do not. We propose \ours, which spends visual resolution on finding and recognizing evidence and language tokens on using it. Without keeping the source text, it doubles the accuracy of a frozen Qwen3-VL-8B on long documents compressed $5\times$ and $10\times$: at $5\times$, \textbf{from 27.7\% to 54.8\%}, near the original text's 58.5\%, with a context smaller than the compressed pages themselves. For less than the cost of reading the original text, it leads at equal budget every method that does without it, at every level, and beats AGAR and VERA, which paste the original text back; at $5\times$ it also beats LensVLM-9B, which is trained and keeps the original text. Visual-text compression is limited not by how much text a model can recover, but by how much it can use.
\end{abstract}
\section{Introduction}
\label{sec:intro}

Rendering text as images has become a way to fit more of it into a vision-language model (VLM). Visual-text compression renders long contexts into pages so that each visual token carries several words~\citep{glyph,deepseekocr}, and document VLMs read scanned pages at whatever resolution the context allows~\citep{pix2struct,qwen3vl,mmlongbenchdoc}. Neither keeps the source text, and both are engineered around one question: \emph{can the model read the text?} Methods that improve on compression make the text readable again: they train the reader to expand pages back into text~\citep{lensvlm}, re-render located regions from the original text~\citep{agar,vera}, or zoom in where the model looks~\citep{vicrop,deepeyes}. The assumption behind all of them is that text a model can read is text it can use.

This assumption fails. Give a VLM only the evidence a question needs, rendered as an image. Asked to transcribe the image, it copies the answer out verbatim; given back its own transcript as text, it answers nearly as well as from the original text; asked the question about the image itself, it is 7--14 points worse (\cref{fig:teaser,app:readers}), across three model families and from 2B to 32B parameters. We call this difference, on evidence the reader recognized, the \emph{\gap}. The question is not whether a model can read, but whether it can use what it reads.

Spending more on the image closes the gap only by giving back the savings. Qwen3-VL-8B recognizes 84\% of the evidence at 15\,px characters, yet its answers from the image catch up only at 24\,px, which costs 2.6$\times$ the visual tokens: the very tokens that compression was meant to save. Nor does scale help: from 2B to 32B, recognition rises from 62\% to 88\%, while the gap stays between 11 and 14 points.

What does close the gap is saying the evidence. The readers with small gaps all think before answering, and 87--88\% of their traces quote the evidence verbatim. Prefill an empty thought, and the gap returns; give them instead their own transcript, written without seeing the question, and they recover most of what thinking adds. Editing their own thinking shows that what matters is the answer in words: a trace cut to the quoted evidence or to the reasoning recovers most of the benefit if it still states the answer, and none if the answer is removed. Prefilled with the correct answer, they adopt it far more often than a wrong entity from the same evidence. These results point to one principle:
\begin{center}
\fcolorbox{gainc}{oursbg}{\begin{minipage}{0.93\linewidth}\centering
\emph{Recognizing text in an image and using it are \textbf{separate abilities} with \textbf{different resolution requirements}; putting the recognized evidence \textbf{into words} closes the gap between them.}
\end{minipage}}
\end{center}

The principle also says how much words should help a reader, and the gap ranks it. In a preregistered held-out test on whole documents, the gap measured on rendered evidence ranks how much eight readers gain from their own words over pixels (Spearman $\rho = 0.86$), while recognition rates do not; and one reader's words lose to pixels on sharp pages, where it has no gap, but win at low resolution.

\ours puts the principle to work. The reader's own attention finds the pages at low resolution; it reads them aloud where recognition saturates and answers from its words alone. Where the found pages are already sharp enough for the reader to use, as with native scans, it hands over the pages themselves. It trains no reader and keeps no original text.

Our contributions are:
\begin{itemize}
\item \textbf{Principle Discovery.} Recognizing and using visual text are separate abilities with different resolution requirements, and putting the evidence into words closes the \gap between them: scale buys only recognition, thinking helps through the answer in words, and the gap ranks, out of sample, how much words beat pixels.
\item \textbf{Practical Solution.} \ours spends visual resolution on locating and recognizing evidence and words on using it, with no source text and no reader training.
\item \textbf{Experimental Evaluation.} \ours is \emph{high-performing}: $+30.9$, $+19.8$ and $+10.9$ points at equal budget under $5\times$, $10\times$ and $15\times$ compression, ahead, for less than the cost of reading the original text, of every method that does without it and, at $5\times$, of LensVLM-9B, which is trained and stores the original text; \emph{transferable}: it lifts Glyph on LongBench~\citep{longbench} and RULER~\citep{ruler}, Gemini~4, a 4B reader and real documents; and \emph{compact}: at $5\times$ it doubles the accuracy from the compressed pages (27.7\% to 54.8\%) from an even smaller context, within 4 points of the original text from a context a seventh as long.
\end{itemize}
\section{Characterizing the Read--Use Gap}
\label{sec:gap}

\begin{figure*}[t]
\centering
\includegraphics[width=\textwidth]{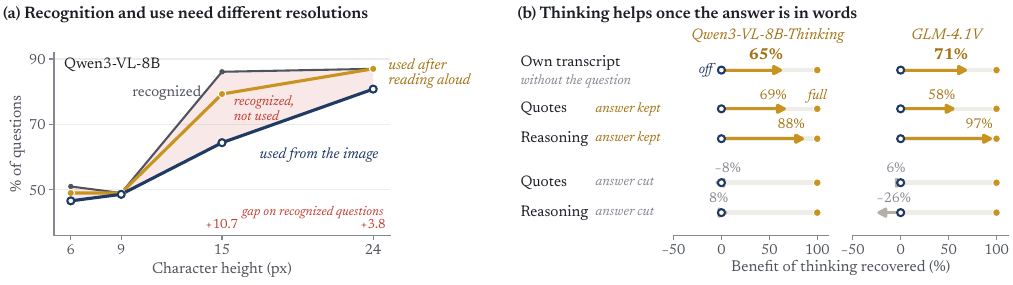}
\caption{\textbf{Recognition and use respond differently to resolution, and thinking helps once the answer is in words.} (a) Qwen3-VL-8B on text-answerable questions: recognition saturates at 15\,px, while answering from the image catches up only at 24\,px; reading aloud tracks recognition. Shaded: recognized but not used, over all questions; red numbers: the gap $G$ on recognized questions. (b) Arrows run from no thinking (0\%) to the share of the full trace's benefit recovered, on recognized questions. Own transcript: the reader answers from its transcript, written without the question, with no thinking. Below, the reader's own trace is cut to the evidence it quotes or to the reasoning around it, with the answer it gave kept or every mention of it removed, and prefilled as its thinking. Whatever keeps the answer in words recovers the benefit; nothing without it does.}
\label{fig:density}
\end{figure*}

\subsection{Recognizing Evidence versus Using It}
\label{sec:gap-setup}

\paragraph{Protocol.}
We isolate reading from search. For each question we take only its gold supporting evidence from long-document QA (HotpotQA~\citep{hotpotqa}, MuSiQue~\citep{musique} and NQ~\citep{nq} in the LensVLM setting; 100 questions each), render it at the density of $10\times$ visual-text compression, with characters about 6\,px tall (we give resolution as character height), and enlarge the same image to 9, 15 and 24\,px, so every size carries the same information and differs only in visual tokens per character. Every reader answers the same 300 images in three ways: from the original evidence text (\emph{Text}); from the image, in one pass (\emph{Image}); and from its own transcript of the image, given as text without the image (\emph{Own transcript}).

\paragraph{Recognition and the gap.}
A question is \emph{recognized} when the gold answer appears verbatim in a second transcript the reader makes of the same image with another prompt, apart from the one it answers from; on recognized questions the evidence is already in the reader's hands, so a failure is a failure of use. For reader $m$ at height $r$, the \gap is
\begin{equation}
G(m,r) = \mathrm{Acc}_{\text{transcript}} - \mathrm{Acc}_{\text{image}},
\end{equation}
both computed on the questions $m$ recognizes at $r$. Accuracy is substring exact match, with 95\% bootstrap CIs.

\paragraph{Readers.}
Instruction-tuned: Qwen3-VL 2B/4B/8B/32B~\citep{qwen3vl}, InternVL3.5-8B~\citep{internvl35} (InternViT encoder) and Gemma-3-12B~\citep{gemma3} (SigLIP encoder). Thinking: GLM-4.1V-9B-Thinking~\citep{glm41v}, Qwen3-VL-8B-Thinking, Glyph (trained from the GLM-4.1V base on rendered text) and Elorian Glimpse. Frontier: GPT-6.1 and Gemini~4 on a 90-question subset.\footnote{API versions gpt-6.1-sol and gemini-4-argon, and Elorian Glimpse through OpenRouter, accessed in October 2026.}

\obs{1}: \textbf{VLMs recognize evidence they cannot use.}
At 15\,px, Qwen3-VL-8B recognizes 84\% of the evidence. On those questions it answers 66.4 from its own transcript, close to 70.8 from the original text, but 55.7 from the image it has just transcribed: a gap of $+10.7$ (CI $[+4.7, +16.6]$). Every instruction-tuned reader answers from its own transcript within 4.4 points of the original text, and every one shows the gap, from $+7.1$ (Gemma) to $+14.0$ (Qwen3-VL-32B; \cref{app:readers}). Frontier readers have smaller gaps ($+3.6$ for GPT-6.1, none for Gemini~4, a thinking reader), as \obs{3} explains. Wrong answers from the image are mostly other entities in the same evidence (71\% appear verbatim in the transcript), not misread strings. The gap is not an artifact of prior knowledge or misreading: it reaches $+15.4$ on synthetic records no prior can answer (\cref{app:alt}), and in the very pass that answers wrongly, the answer's words are as readable on the visual tokens as when it answers right (\cref{app:mech}).

\subsection{Different Resolution Requirements}
\label{sec:gap-density}

\obs{2}: \textbf{Recognizing visual text needs fewer pixels than using it.}
\Cref{fig:density}a sweeps character height for Qwen3-VL-8B on questions it answers correctly from text. Recognition jumps from 49\% to 86\% between 9 and 15\,px and then saturates; answering from the image is at 64\% at 15\,px and reaches 81\% only at 24\,px, which costs 2.6$\times$ the visual tokens, while its own transcript already gives 79\%. On recognized questions the gap is still $+10.7$ at 15\,px and falls to $+3.8$ only at 24\,px. Extra pixels buy recognition first and use later; reading aloud buys use with about 200 text tokens.

\textbf{Scale does not close the gap, and real documents obey the same law.} Larger readers recognize more but use no better: from Qwen3-VL-2B to 32B, recognition at 15\,px rises from 62\% to 88\% while the gap stays at $+11.3$, $+13.9$, $+10.7$ and $+14.0$, still significant for 32B at 24\,px ($+5.8$) (all readers in \cref{fig:grid}). On MMLongBench-Doc evidence pages, Qwen3-VL-8B has no gap at the native 144\,DPI, while at 48\,DPI the gap is $+25.8$ on text evidence and also opens on tables and charts; on purely pictorial evidence, with nothing to recite, transcribing hurts ($-8.3$; \cref{tab:app-docpages}). Whether converting a page to text helps depends on resolution and on how much of the evidence is text, which explains why OCR-first pipelines sometimes help and sometimes hurt~\citep{rest}.

\begin{figure}[t]
\centering
\includegraphics[width=\columnwidth]{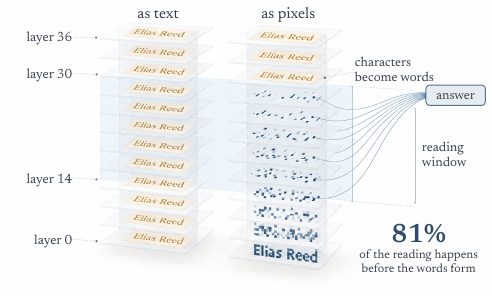}
\caption{\textbf{The answer reads the image before its characters become words.} Schematic of Qwen3-VL-8B; the layers and the 81\% are measured. Shaded: the layers in which the answer reads the evidence, found by cutting its attention to the evidence layer by layer (pixels 14--30, text 12--28). Under the logit lens, the words drawn in the image become decodable only from layer 24, reaching half their peak at layer 30.}
\label{fig:mech}
\end{figure}

\subsection{Recitation in Reasoning Traces}
\label{sec:gap-think}

\obs{3}: \textbf{Thinking closes the gap by putting the answer into words.}
The readers with small gaps all think before answering, and their thinking quotes the evidence: 87\% of GLM-4.1V's and 88\% of Qwen3-VL-8B-Thinking's traces share a run of at least six consecutive words with the reader's own transcript. Prefilling an empty thought brings the gap back to the level of instruction-tuned readers, from $-1.5$ to $+11.2$ and from $+6.0$ to $+13.1$ (\cref{fig:teaser,app:think}). What thinking supplies is words, not reasoning about the question: the reader's own transcript, written without seeing the question, recovers most of what thinking adds (71\% and 65\%). Editing each reader's own traces shows that the benefit is carried by the answer in words (\cref{fig:density}b): a trace cut to the quoted evidence or to the reasoning recovers most of the benefit of thinking if it still states the answer (58--97\%), and none if the answer is removed ($-26\%$ to $8\%$, none significant). The reader checks the words against the image: prefilled as its thinking, the correct answer is adopted 90\% and 94\% of the time, a wrong entity from the same evidence only 23\% and 37\%, and the correct answer is adopted most of the time even where its own thinking got the question wrong (\cref{app:cf}). The reader recognizes the answer once it is in words, yet does not put it into words on its own. Instruction-tuned readers fail partly the same way: stating the answer among three other entities from the same evidence halves their gap ($+16.7$ to $+8.6$; \cref{app:cf}), and the rest closes only with the evidence itself in words. Glyph's lack of a gap needs no rendered-text training: GLM-4.1V-9B-Thinking, its base, recites the evidence the same way and has none either.

\subsection{Inside the Reader}
\label{sec:gap-mech}

\obs{4}: \textbf{The answer reads the image before its characters become words.}
Cutting attention from the question and answer to the evidence, layer by layer~\citep{geva}, shows when the answer reads it (\cref{fig:mech}): in Qwen3-VL-8B, from layer 14 to 30 of 36 for enlarged pixels and from 12 to 28 for text, the same window in both forms. The words drawn in the image become decodable under the logit lens~\citep{logitlens} only from layer 24 and reach half their peak at layer 30, so 81\% of what the answer takes from the image is taken before its characters are words. Depth moves both together: at every scale the window closes 6--8 layers before the output, and most of the reading precedes the words at every scale (61--99\%; \cref{app:mech}). Text enters as words, so evidence read aloud is already words in the layers where the answer reads it.

\paragraph{Summary.}
What closes the gap is the words of the evidence, whether the reader transcribes it on request or recites it in its own thinking, and it is the words, not the extra tokens: prefilling its own transcript beats 200 dots by 6.1 points and another question's transcript by 12.2 (\cref{app:alt}).
\section{The Gap Ranks How Much Words Beat Pixels}
\label{sec:predict}

If recognition and use are separate abilities and words close the gap, the gap should predict how much a reader gains from its own words over pixels. We test this out of sample, on another corpus and task, with registered predictions.

\paragraph{Test bed.}
MMLongBench-Doc~\citep{mmlongbenchdoc} whole documents: 745 answerable questions, 47 pages on average. Every reader sees the whole document at 48\,DPI; the scan of \cref{sec:method} picks two pages, and the reader gets a second look at them as \emph{pixels} (the page images) or as \emph{words} (its own transcript of them), at 144\,DPI or at the same 48\,DPI. The \emph{representation benefit} is $\Delta = \mathrm{Acc}(\text{words}) - \mathrm{Acc}(\text{pixels})$ on the 576 text-evidence questions, judged by an LLM (DeepSeek V4.1 Flash).

\paragraph{Predictors.}
We register two predictors (\cref{app:prereg}), both measured on rendered long-document evidence: the 15\,px gap on the 76 questions every reader recognizes, to rank readers at 144\,DPI; and the gap at the character height matched to the second look, $G(m, 24\,\text{px})$ for 144\,DPI and $G(m, 6\,\text{px})$ for 48\,DPI. Eight readers enter: the six instruction-tuned readers of \cref{sec:gap-setup}, GLM-4.1V-9B-Thinking and Glyph.

\begin{figure}[t]
\centering
\includegraphics[width=\linewidth]{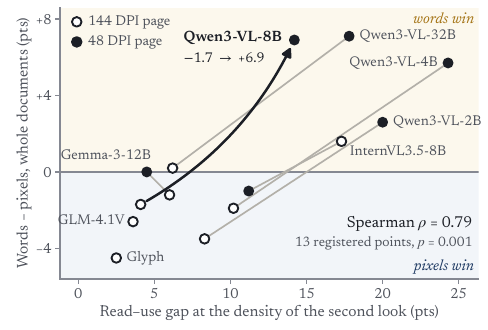}
\caption{\textbf{The read--use gap ranks the advantage of words over pixels} across readers and resolutions. $x$: gap measured on rendered long-document evidence at the density of the second look. $y$: words minus pixels on MMLongBench-Doc whole documents. Lines join the same reader at the two densities. Spearman $\rho$ over the 13 registered points; Qwen3-VL-8B at 144\,DPI, observed before this predictor was registered, is drawn but not counted.}
\label{fig:predict}
\end{figure}

\obs{5}: \textbf{The read--use gap ranks how much words beat pixels, beyond what recognition explains.}
At 144\,DPI, the 15\,px gap ranks the representation benefit of the eight readers with Spearman $\rho = 0.86$ (exact one-sided $p = 0.005$), within the Qwen family ($0.95$) and among the other four readers ($1.00$; together, exact $p = 0.003$; \cref{tab:app-rank}). Across both densities, the density-matched gap predicts the benefit over all 13 registered points with $\rho = 0.79$ (permutation $p = 0.001$; \cref{fig:predict}), and still within each density (stratified $p = 0.035$). Recognition rates do not ($\rho = 0.38$ at 144\,DPI, $-0.53$ across both densities). The benefit flips with density within a reader: Qwen3-VL-8B's words trail its pixels by 1.7 points at 144\,DPI and lead by 6.9 at 48\,DPI (CI $[+4.0, +9.9]$). GLM-4.1V, which already recites the evidence in its thinking, gains nothing from words at either density.
\section{\texorpdfstring{\ours}{ALOUD}}
\label{sec:method}

The read--use gap shows that finding evidence, recognizing it and using it place different demands on a reader. Rendering a whole document at uniformly high resolution meets all three with the same expensive resource, image tokens. \ours separates them: a low-resolution scan finds the evidence, and reading it aloud turns what the reader recognizes into words it can use. Both steps use the reader's own abilities: \ours trains no reader and needs no original text; its one fitted part is a linear readout on the reader's attention.

\begin{figure*}[t]
\centering
\includegraphics[width=\textwidth]{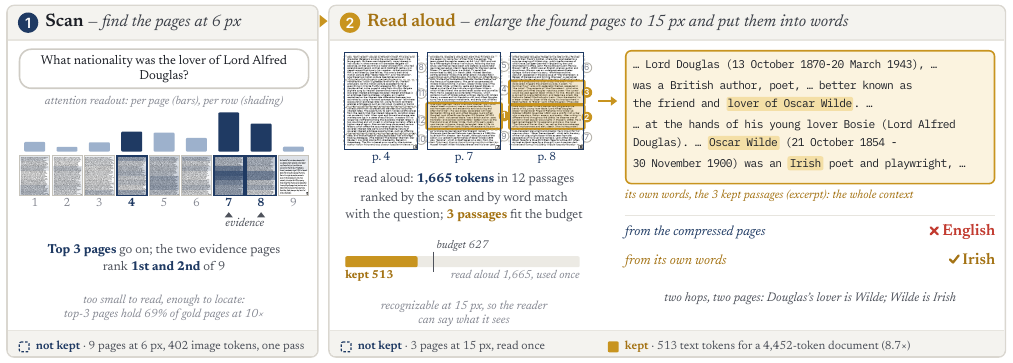}
\caption{\textbf{\ours on a real question} (HotpotQA, $10\times$ compression, nine pages). \textcircled{\scriptsize 1}~\textbf{Scan}: in one low-resolution pass, an attention readout scores every row of visual tokens; a page scores as its best row, and the top three pages go on. Both evidence pages are found, though too small to read. \textcircled{\scriptsize 2}~\textbf{Read aloud}: the reader transcribes the three pages enlarged to 15\,px; its 12 passages are ranked by the scan and by their word match with the question, and the three that fit the budget become the whole context. From the compressed pages the reader answers English; from its own words, Irish. Neither the compressed nor the enlarged pages stay in the context (513 tokens for a 4,452-token document).}
\label{fig:method}
\end{figure*}

\subsection{Scan: Locate Pages at Low Resolution}
The reader reads the whole compressed document and the question in one forward pass. For every row of visual tokens we pool the attention that the question tokens pay to it, summed within the row, per layer and head. A logistic readout fitted on calibration questions turns these features into a row score, and a page scores as its best row (\cref{app:impl}). The reader knows where the answer is long before it can read it (\cref{sec:eval}), so locating costs only low-resolution tokens.

\subsection{Read Aloud: Put the Found Pages into Words}
The top pages are enlarged, by bicubic interpolation of the compressed pages themselves, until characters are about 15\,px tall, where recognition saturates (\cref{sec:gap-density}), and the reader transcribes each of them. Its words then replace the pages: the reader answers from its own transcript alone. To fit the budget, the transcript is cut into passages of about 100 words, ranked by fusing the scan score of the rows each passage covers with its BM25 match to the question (reciprocal rank fusion), and kept in that order up to the token budget BM25 gets in \cref{tab:main} (\cref{app:impl}); \ours fills 73--85\% of it. The number of pages and the ranking are chosen once on calibration questions: three pages with the fused ranking at $5\times$ and $10\times$, four by BM25 alone at $15\times$, where the scan is blurriest.

The principle also says when words are not needed: answering from the image catches up with reading aloud at 24\,px for our reader (\cref{fig:density}a), and a transcript drops layout and charts. \ours therefore hands over the found pages when their native resolution reaches that height, as with scans, and reads them aloud otherwise; the height is a property of the rendering or scan (\cref{sec:predict}).
\section{Evaluation}
\label{sec:eval}

We ask three questions. \textbf{RQ1}: Can a low-resolution scan locate the evidence? \textbf{RQ2}: Does \ours improve question answering over compressed documents? \textbf{RQ3}: Does it carry over to a reader trained on rendered text, to real documents and to frontier models?

\subsection{Setup}
The reader is a frozen Qwen3-VL-8B. RQ1 and RQ2 follow the LensVLM evaluation: HotpotQA, MuSiQue and NQ documents of 3K--12K tokens rendered at its $5\times$, $10\times$ and $15\times$ presets, 200 test and 400 calibration questions per dataset, substring exact match. We budget what the reader answers from. The \emph{effective compression rate} is the original-text tokens over the tokens of the context the answer is generated from, transcripts and thinking included (for LensVLM, its single growing context). Passes whose inputs are gone before the answer, such as our scan and reading aloud and ViCrop's reference pass, count as compute, as the pass that chooses what to keep does in prompt and KV-cache compression~\citep{llmlingua,snapkv,kvzip}; \obs{10} reports it. These discarded passes are not budgeted but are capped: together they must cost less than reading the original text, beyond which the compressed context saves nothing. The \emph{same-budget gain} is accuracy minus that of uniform rendering at the same rate, interpolated linearly between nine densities (0.6--18$\times$).

Baselines run with their official prompts and protocols. \emph{Transcribe all}: the same reader, or DeepSeek-OCR~\citep{deepseekocr}, transcribes every page and the reader answers from the text. \emph{Think first}: Qwen3-VL-8B-Thinking reads the same pages (recommended sampling, 24{,}576-token limit). \emph{Zoom tool}: a zero-shot zoom agent, the same reader allowed to ask for up to two pages enlarged, and DeepEyes~\citep{deepeyes}, trained with reinforcement learning to zoom, which takes the pages tiled into a single image. \emph{Attention crop}: ViCrop~\citep{vicrop} and VERA~\citep{vera} with pixels. \emph{Trained}: Glyph~\citep{glyph}. For reference, methods that need the original text, which this setting does not store: AGAR~\citep{agar}, VERA with text, LensVLM-9B~\citep{lensvlm}, trained to expand pages back into their stored text, and BM25~\citep{bm25} over the original text with the budget of \ours; and two methods above the cap, which transcribe every page and search the transcript with BM25 within the same budget: our reader, which is \ours without the scan, and DeepSeek-OCR. Their transcripts alone run to 9.4K--26K tokens against the text's 8.8K, while the passes \ours discards take 3.7K--4.0K.

\subsection{Locating Evidence (RQ1)}
\obs{6}: \textbf{The reader finds the evidence long before it can read it.}
In one low-resolution pass, the pages it reads (three, four at $15\times$) cover 75\%, 69\% and 45\% of the gold pages at $5\times$, $10\times$ and $15\times$ (random: 16\%, 21\%, 20\%), and 77\% on whole MMLongBench-Doc documents at 48\,DPI, while at $10\times$ the reader recognizes only 47\% of the evidence. Attention localizes pages rather than lines, so \ours reads whole pages.

\subsection{QA over Compressed Documents (RQ2)}
\begin{table*}[t]
\caption{\textbf{Compressed long documents.} Same-budget gain (points) over uniform rendering for a frozen Qwen3-VL-8B in the LensVLM setting (HQA: HotpotQA; MSQ: MuSiQue). Everything left in the answering context counts, transcripts and thinking included; passes whose outputs are gone before the answer count as compute (\cref{tab:app-cost}). Listed for reference and not ranked: methods that need the original text, and methods above the cost cap of \cref{sec:eval}. $^\dagger$Trains the reader. \ours reads aloud the pages its scan finds (three; four at $15\times$); \ours (1 page) only the top one. \textbf{Bold}: above every ranked baseline; \underline{underline}: strongest ranked baseline; \gain{x}: gain over it.}
\label{tab:main}
\centering\footnotesize
\setlength{\tabcolsep}{1.3pt}
\setlength{\aboverulesep}{0pt}\setlength{\belowrulesep}{0pt}
\renewcommand{\arraystretch}{1.12}
\begin{tabular}{p{1.95cm}l>{\hspace{1pt}}rrrr<{\hspace{1pt}}>{\hspace{1pt}}rrrr<{\hspace{1pt}}>{\hspace{1pt}}rrrr<{\hspace{1pt}}}
\toprule
\headrow & & \multicolumn{4}{c}{\textbf{5$\times$}} & \multicolumn{4}{c}{\textbf{10$\times$}} & \multicolumn{4}{c}{\textbf{15$\times$}} \\
\cmidrule(lr){3-6}\cmidrule(lr){7-10}\cmidrule(lr){11-14}
\headrow \textbf{Group} & \textbf{Method} & \multicolumn{1}{c}{\textbf{HQA}} & \multicolumn{1}{c}{\textbf{MSQ}} & \multicolumn{1}{c}{\textbf{NQ}} & \multicolumn{1}{c}{\textbf{Avg.}\hspace*{0.6\gainw}} & \multicolumn{1}{c}{\textbf{HQA}} & \multicolumn{1}{c}{\textbf{MSQ}} & \multicolumn{1}{c}{\textbf{NQ}} & \multicolumn{1}{c}{\textbf{Avg.}\hspace*{0.6\gainw}} & \multicolumn{1}{c}{\textbf{HQA}} & \multicolumn{1}{c}{\textbf{MSQ}} & \multicolumn{1}{c}{\textbf{NQ}} & \multicolumn{1}{c}{\textbf{Avg.}\hspace*{0.6\gainw}} \\
\midrule
 & \textcolor{demph}{AGAR} & \textcolor{demph}{9.5} & \textcolor{demph}{2.6} & \textcolor{demph}{1.3} & \textcolor{demph}{4.5}\gslot{} & \textcolor{demph}{1.3} & \textcolor{demph}{$-$1.6} & \textcolor{demph}{3.5} & \textcolor{demph}{1.1}\gslot{} & \textcolor{demph}{$-$0.7} & \textcolor{demph}{1.2} & \textcolor{demph}{2.1} & \textcolor{demph}{0.9}\gslot{} \\  %
 & \textcolor{demph}{VERA (text)} & \textcolor{demph}{22.4} & \textcolor{demph}{11.4} & \textcolor{demph}{15.3} & \textcolor{demph}{16.4}\gslot{} & \textcolor{demph}{12.6} & \textcolor{demph}{5.0} & \textcolor{demph}{13.4} & \textcolor{demph}{10.3}\gslot{} & \textcolor{demph}{$-$1.3} & \textcolor{demph}{$-$2.7} & \textcolor{demph}{9.3} & \textcolor{demph}{1.8}\gslot{} \\  %
 & \textcolor{demph}{LensVLM-9B$^\dagger$} & \textcolor{demph}{31.1} & \textcolor{demph}{20.4} & \textcolor{demph}{22.0} & \textcolor{demph}{24.5}\gslot{} & \textcolor{demph}{28.4} & \textcolor{demph}{14.8} & \textcolor{demph}{27.2} & \textcolor{demph}{23.5}\gslot{} & \textcolor{demph}{18.6} & \textcolor{demph}{8.0} & \textcolor{demph}{17.1} & \textcolor{demph}{14.5}\gslot{} \\  %
\multirow{-4}{1.95cm}{\raggedright\itshape \textcolor{demph}{Original text}} & \textcolor{demph}{BM25} & \textcolor{demph}{34.5} & \textcolor{demph}{19.8} & \textcolor{demph}{28.9} & \textcolor{demph}{27.8}\gslot{} & \textcolor{demph}{40.7} & \textcolor{demph}{22.3} & \textcolor{demph}{33.7} & \textcolor{demph}{32.2}\gslot{} & \textcolor{demph}{41.3} & \textcolor{demph}{18.7} & \textcolor{demph}{35.2} & \textcolor{demph}{31.7}\gslot{} \\  %
\midrule
 & \textcolor{demph}{\ours, every page} & \textcolor{demph}{34.1} & \textcolor{demph}{18.0} & \textcolor{demph}{27.3} & \textcolor{demph}{26.5}\gslot{} & \textcolor{demph}{34.2} & \textcolor{demph}{12.4} & \textcolor{demph}{23.0} & \textcolor{demph}{23.2}\gslot{} & \textcolor{demph}{22.4} & \textcolor{demph}{6.9} & \textcolor{demph}{21.2} & \textcolor{demph}{16.8}\gslot{} \\  %
\multirow{-2}{1.95cm}{\raggedright\itshape \textcolor{demph}{Costlier than the text}} & \textcolor{demph}{DeepSeek-OCR + BM25} & \textcolor{demph}{32.0} & \textcolor{demph}{21.3} & \textcolor{demph}{30.2} & \textcolor{demph}{27.8}\gslot{} & \textcolor{demph}{27.6} & \textcolor{demph}{12.9} & \textcolor{demph}{29.1} & \textcolor{demph}{23.2}\gslot{} & \textcolor{demph}{2.9} & \textcolor{demph}{3.8} & \textcolor{demph}{6.6} & \textcolor{demph}{4.4}\gslot{} \\  %
\midrule
 & Same reader & 8.7 & 4.5 & 3.3 & 5.5\gslot{} & 1.7 & $-$2.0 & $-$4.6 & $-$1.6\gslot{} & $-$10.5 & $-$13.2 & $-$15.0 & $-$12.9\gslot{} \\  %
\multirow{-2}{1.95cm}{\raggedright\itshape Transcribe all} & DeepSeek-OCR & 4.1 & 7.0 & 4.4 & 5.1\gslot{} & $-$7.0 & $-$7.0 & 2.0 & $-$4.0\gslot{} & $-$32.0 & $-$21.0 & $-$24.5 & $-$25.8\gslot{} \\  %
\midrule
\multirow{-1}{1.95cm}{\raggedright\itshape Think first} & Thinking reader & $-$33.8 & $-$26.0 & $-$32.5 & $-$30.8\gslot{} & $-$25.8 & $-$19.2 & $-$19.0 & $-$21.3\gslot{} & $-$10.7 & $-$16.1 & $-$13.5 & $-$13.4\gslot{} \\  %
\midrule
 & Zoom agent & $-$3.8 & $-$1.3 & 3.9 & $-$0.4\gslot{} & 3.4 & 1.1 & 5.4 & 3.3\gslot{} & 1.4 & 2.4 & $-$0.8 & 1.0\gslot{} \\  %
\multirow{-2}{1.95cm}{\raggedright\itshape Zoom tool} & DeepEyes$^\dagger$ & $-$19.1 & $-$22.7 & $-$22.0 & $-$21.2\gslot{} & $-$32.0 & $-$24.7 & $-$29.2 & $-$28.6\gslot{} & $-$32.4 & $-$25.3 & $-$39.3 & $-$32.4\gslot{} \\  %
\midrule
 & ViCrop & 9.2 & 8.6 & 8.7 & 8.9\gslot{} & 4.6 & $-$0.1 & 5.3 & 3.3\gslot{} & 0.5 & 1.0 & 1.2 & 0.9\gslot{} \\  %
\multirow{-2}{1.95cm}{\raggedright\itshape Attention crop} & VERA (pixels) & \underline{14.4} & \underline{9.2} & \underline{14.4} & \underline{12.7}\gslot{} & 8.9 & 1.3 & \underline{6.5} & 5.6\gslot{} & $-$0.7 & 0.8 & 1.4 & 0.5\gslot{} \\  %
\midrule
\multirow{-1}{1.95cm}{\raggedright\itshape Trained} & Glyph$^\dagger$ & 7.6 & 7.2 & 7.2 & 7.3\gslot{} & \underline{9.5} & \underline{7.3} & 2.9 & \underline{6.6}\gslot{} & \underline{7.7} & \underline{6.7} & \underline{4.1} & \underline{6.2}\gslot{} \\  %
\midrule
\oursrow  & \ours (1 page) & \textbf{26.5} & \textbf{15.0} & \textbf{36.0} & \textbf{25.8}\gslot{\gain{13.1}} & \textbf{14.0} & 5.4 & \textbf{23.7} & \textbf{14.4}\gslot{\gain{7.8}} & 5.5 & 3.5 & \textbf{12.0} & \textbf{7.0}\gslot{\gain{0.8}} \\  %
\oursrow \multirow{-2}{1.95cm}{\raggedright\itshape Ours} & \textbf{\ours} & \textbf{37.9} & \textbf{22.6} & \textbf{32.3} & \textbf{30.9}\gslot{\gain{18.2}} & \textbf{23.6} & \textbf{8.7} & \textbf{27.2} & \textbf{19.8}\gslot{\gain{13.2}} & \textbf{10.0} & 4.4 & \textbf{18.4} & \textbf{10.9}\gslot{\gain{4.7}} \\  %
\bottomrule
\end{tabular}
\end{table*}

\begin{figure}[t]
\centering
\includegraphics[width=\linewidth]{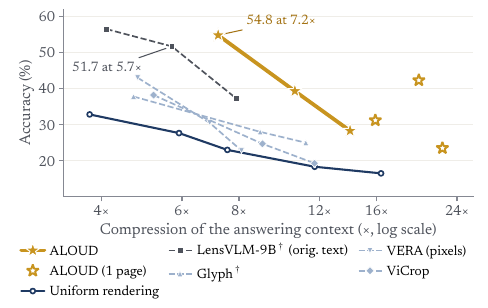}
\caption{\textbf{Accuracy against the compression of the answering context} (original-text tokens over the tokens the reader answers from), mean over the three datasets; one point per compression level. Methods whose answering context exceeds a third of the original text are off the axis (\cref{app:abs}). $^\dagger$Trains the reader.}
\label{fig:pareto}
\end{figure}

\obs{7}: \textbf{Without training or original text, \ours is the strongest reader of compressed documents.}
At $5\times$, $10\times$ and $15\times$, \ours gains $+30.9$, $+19.8$ and $+10.9$ points at equal budget and leads every method that does without the original text for less than the cost of reading it, at every level (\cref{tab:main}): by 23.6, 13.2 and 4.7 points over Glyph, trained on rendered text, and by 10--18 over VERA with pixels; it also leads AGAR and VERA with the original text pasted back (all paired CIs exclude zero). At $5\times$ it beats LensVLM-9B, trained and storing the original text, by $+6.4$ at equal budget (CI $[+1.9, +10.8]$). Compared directly (\cref{fig:pareto}), \ours answers 54.8 at 7.2$\times$, above LensVLM at 5.7$\times$ (51.7), against 27.7 from the pages alone at 5.9$\times$ and 58.5 from the original text; reading only the top page, it matches the strongest ranked baseline (42.3 vs.\ 43.0) from a context four times smaller. Words make the found evidence usable: the same passages as pixels enlarged to 15\,px take 40--53\% of the tokens but answer 4.8--10.8 points worse, and still 3.5--8.5 worse when the pixels fill the whole budget (\cref{app:ablation}). Pixels are the cheaper way to store evidence; words are the better way to use it. The baselines stop where the principle says: attention crops and zoom agents find the evidence but leave it in pixels. Reading aloud is the gain; the scan is the compression. Read aloud and searched within the budget, every page already beats every method that leaves the evidence in pixels ($+26.5$, $+23.2$, $+16.8$) but costs more than the original text; the scan cuts the words five- to sevenfold and keeps most of the gain, exceeding it at $5\times$. DeepSeek-OCR in its place ties at $10\times$ ($+23.2$) and fails at $15\times$ ($+4.4$) on characters our reader still reads by enlarging them to 15\,px. Thinking readers recite too, but inside the answering context, at the budget's expense; \ours recites outside it and keeps only what the answer needs.

\subsection{Transfer (RQ3)}
\begin{table}[t]
\caption{\textbf{\ours carries over} to a reader trained on rendered text (Glyph, on its own protocol), to real documents and to Gemini~4. $+$\,\ours: with the evidence a frozen Qwen3-VL-8B scanned and read aloud (W), or, on scanned pages, the enlarged pages themselves (P; \cref{sec:method}); both forms in \cref{app:forms}. MMLongBench-Doc: uniform 72\,DPI vs.\ 48\,DPI plus two pages at 144\,DPI. Gemini~4: 600 questions; GPT-6.1: 60. \gain{x}: gain over Alone.}
\label{tab:transfer}
\centering\footnotesize
\setlength{\tabcolsep}{2.2pt}
\setlength{\aboverulesep}{0pt}\setlength{\belowrulesep}{0pt}
\renewcommand{\arraystretch}{1.12}
\begin{tabular}{llcrr}
\toprule
\headrow \textbf{Benchmark} & \textbf{Reader} & \textbf{As} & \multicolumn{1}{c}{\textbf{Alone}} & \multicolumn{1}{c}{\textbf{$+$\,\ours}\hspace*{0.6\gainw}} \\
\midrule
RULER, DPI 72 & Glyph & W & 73.2 & \textbf{79.8}\gslot{\gain{6.6}} \\  %
LongBench, DPI 72 & Glyph & W & 53.4 & \textbf{60.1}\gslot{\gain{6.7}} \\  %
LongBench, DPI 48 & Glyph & W & 39.4 & \textbf{51.7}\gslot{\gain{12.3}} \\  %
MMLongBench-Doc & Qwen3-VL-8B & P & 39.7 & \textbf{45.6}\gslot{\gain{5.9}} \\  %
Long-doc QA, $10\times$ & GPT-6.1 & W & 52.5 & 55.0\gslot{} \\  %
Long-doc QA, $10\times$ & Gemini 4 & W & 75.7 & \textbf{82.3}\gslot{\gain{6.6}} \\  %
\bottomrule
\end{tabular}
\end{table}

\obs{8}: \textbf{The gains carry over to other readers, benchmarks and real documents.}
\textbf{A reader trained on rendered text.} With Glyph's own rendering and protocol, our readout unchanged, the evidence our frozen 8B reader read aloud lifts Glyph on RULER at every length up to 126K (\cref{app:forms}) and on LongBench QA at DPI 72 and 48 (\cref{tab:transfer}).
\textbf{Real documents.} On whole MMLongBench-Doc documents, the two scanned pages at native 144\,DPI beat uniform 72\,DPI by $+5.9$ (CI $[+3.0, +8.9]$) with 64\% of its tokens.
\textbf{Other readers.} With its own scan and transcripts, Qwen3-VL-4B gains $+30.4$, $+18.7$ and $+5.8$, within 1.1 points of the 8B reader at $5\times$ and $10\times$. Our reader's words lift Gemini~4 by $+6.6$ (600 questions; CI $[+3.5, +9.8]$).

\obs{9}: \textbf{The principle also marks where words stop helping.}
At $15\times$ the reader no longer recognizes the characters reliably (\cref{app:ablation}), and the advantage of its words over the same passages in pixels falls from 10.8 points at $5\times$ to 4.8: words carry only what was recognized. On native 144\,DPI scans, Qwen3-VL-8B has no gap and the pages carry layout and charts, so the pages beat its words (45.6 vs.\ 42.7; \cref{app:forms}).

\subsection{Cost and Ablations}
\obs{10}: \textbf{\ours keeps every pass compressed and spends less than looking harder.}
Its longest pass holds at most a fifth of the original text, and nearly all its compute is reading aloud: about 1{,}800 generated tokens per question at $10\times$; transcribed as four-line strips in one batch, with unchanged accuracy, it answers in about 2\,s on one GPU (\cref{app:cost}). %
Looking harder costs more for less. More pixels: uniform rendering answers 48.5 from 5.4K image tokens, more than the 5.3K \ours processes and generates in all at $5\times$ for 54.8, and needs 13.3K, more than the original text, to reach 52.7. %
More thinking: the thinking reader generates 3{,}800 tokens for 22.2 against our 39.3 at $10\times$. A zoom tool: on 60 questions at $10\times$, Gemini~4 with its own zoom recovers to 79.3 at 6{,}680 billed tokens; given our 8B reader's words, it answers 81.7 in one call at 2{,}015, as well as from the original text. Reading only the top page cuts the generated tokens to about 600, at 16--22$\times$ compression, where no ranked method comes close (\cref{fig:pareto}).
\section{Related Work}
\label{sec:related}

\paragraph{Visual-text compression and document reading.}
VisInContext~\citep{visincontext}, Glyph~\citep{glyph}, DeepSeek-OCR~\citep{deepseekocr} and VIST~\citep{vist} render long text into images, often saving about half the decoder tokens~\citep{textorpixels}, and VTCBench~\citep{vtcbench} tests them. Pixel-based language models~\citep{pixel} and OCR-free document readers~\citep{donut,pix2struct,docowl} learn reading end to end, and current VLMs~\citep{qwen3vl,internvl35,gemma3,glm41v} read whole long documents~\citep{mmlongbenchdoc,longdocurl}. The yardstick is recovery, measured as OCR accuracy or reconstruction at each compression rate~\citep{deepseekocr,opticalae}; recent evaluation sets downstream accuracy aside as confounded by language priors~\citep{zerosenseeval}, although DeepSeek-OCR answers near chance on pages it transcribes well~\citep{visualmerit}. Follow-ups recover lost information by training a model to expand pages back into text (LensVLM~\citep{lensvlm}), a trained localizer (FocusVTC~\citep{focusvtc}), self-distillation (CAPS~\citep{caps}), aligning the image path to the text path (SPIRAL~\citep{spiral}), retrieving the source text of selected pages (SEER~\citep{seer}), or an external OCR beside low-resolution pages (DocVLM~\citep{docvlm}). LensVLM's own tool study finds expanding into text better than expanding into a sharper image on rendered text, and the reverse on native documents; the read--use gap accounts for both. We show that recovered text overstates usable text, and read the evidence aloud with the reader's own abilities, without the original text or training.

\paragraph{Looking closer: zoom, crops and attention.}
VLMs attend to the right region before they can read it~\citep{vicrop}, and zoom tools re-read regions at higher resolution, without training (V*~\citep{vstar}, ZoomEye~\citep{zoomeye}, DyFo~\citep{dyfo}, TextCoT~\citep{textcot}) or with training (Chain-of-Focus~\citep{chainoffocus}, DeepEyes~\citep{deepeyes}, InSight-doc~\citep{insightdoc}, LookZoom~\citep{lookzoom}). On rendered text, AGAR~\citep{agar} and VERA~\citep{vera} read attention heads to find the lines, then re-render them or paste them back from the original text. All of these fetch a sharper view of the source; under compression there is none, and \ours enlarges the pages its scan found and has the reader read them aloud, with no agent deciding when to look again.

\paragraph{Evidence use and modality gaps.}
VLMs answer worse from rendered text than from the same text across more than 30 models (VISTA-Bench~\citep{vistabench}), most on reasoning-heavy tasks (PixelWorld~\citep{pixelworld}, CrossMath~\citep{crossmath}), and on multi-page inputs nearly unchanged from 72 to 300\,PPI (ReadBench~\citep{readbench}), as expected once recognition saturates. SimpleOCR~\citep{simpleocr} finds a capability--utilization gap when the question itself is rendered and trains it away; Reading, Not Thinking~\citep{readingnotthinking} attributes the loss to reduced reasoning, whereas here free-form reasoning~\citep{cot,deepseekr1} recovers little and reciting the evidence closes the gap. IsoBench~\citep{isobench} finds that transcribing diagrams first recovers part of their loss, PRISM-Bench~\citep{prismbench} that self-written descriptions lose spatial evidence, and REST~\citep{rest} that OCR-first pipelines sometimes help and sometimes hurt; we explain when, and rank readers by how much. VLMs still read targets they fail to use after a change of image tiling, and their own transcriptions recover the answers~\citep{readingright}; concurrently, Found but Not Read~\citep{foundnotread} finds that extracted text closes much of this gap and that a crop of the evidence recovers most of it while a box does not, as the principle predicts, and Loud and Clear~\citep{loudclear} finds evidence on retrieved pages underused. CAPS~\citep{caps} reports recovered content that goes unused, and Transcribe, Then Reason~\citep{transcribe} blames a shared output budget. On general images, VLMs often encode evidence they do not use~\citep{encodedunused,bedbench} and use the visual content they can name~\citep{vlmsneedwords}; rendered text is the case where the name is the content. In text, models underuse evidence in long inputs~\citep{lostmiddle} and reciting it helps~\citep{ctxlen,recite}; self-generated hints in VLMs supply knowledge~\citep{lookrecite}, not text read from the image. Attention knockout~\citep{geva,zhangflow} and the logit lens~\citep{logitlens,tunedlens,neo} find visual representations aligning with text late~\citep{nikankin}; on rendered text, the answer reads most of the evidence before then (\cref{fig:mech}). These works report instances of unused content; we explain them, predict out of sample which readers gain from words, and turn the principle into a compression method.
\section{Conclusion}
\label{sec:conclusion}

VLMs recognize more of the text in an image than they can use. The read--use gap is set by resolution, does not close with scale, closes when the evidence is put into words, and ranks how much words beat pixels. \ours applies this principle: visual resolution to find and recognize evidence, words to use it. Without training the reader or storing the source text, it doubles a frozen reader's accuracy on long documents compressed $5\times$ and $10\times$. Visual-text compression has been judged by how much text a model can recover; it should be judged by how much it can use.

{
    \small
    \bibliographystyle{ieeenat_fullname}
    \bibliography{refs}

\begin{thebibliography}{73}
\providecommand{\natexlab}[1]{#1}
\providecommand{\url}[1]{\texttt{#1}}
\expandafter\ifx\csname urlstyle\endcsname\relax
  \providecommand{\doi}[1]{doi: #1}\else
  \providecommand{\doi}{doi: \begingroup \urlstyle{rm}\Url}\fi

\bibitem[Bai et~al.(2025)Bai, Cai, Chen, Chen, Chen, et~al.]{qwen3vl}
Shuai Bai, Yuxuan Cai, Ruizhe Chen, Keqin Chen, Xionghui Chen, et~al.
\newblock {Qwen3-VL} technical report.
\newblock \emph{arXiv preprint arXiv:2511.21631}, 2025.

\bibitem[Bai et~al.(2024)Bai, Lv, Zhang, Lyu, Tang, et~al.]{longbench}
Yushi Bai, Xin Lv, Jiajie Zhang, Hongchang Lyu, Jiankai Tang, et~al.
\newblock {LongBench}: A bilingual, multitask benchmark for long context understanding.
\newblock In \emph{Proceedings of the 62nd Annual Meeting of the Association for Computational Linguistics (ACL)}, 2024.
\newblock arXiv:2308.14508.

\bibitem[Belrose et~al.(2023)Belrose, Ostrovsky, McKinney, Furman, Smith, et~al.]{tunedlens}
Nora Belrose, Igor Ostrovsky, Lev McKinney, Zach Furman, Logan Smith, et~al.
\newblock Eliciting latent predictions from transformers with the tuned lens.
\newblock \emph{arXiv preprint arXiv:2303.08112}, 2023.

\bibitem[Bhatia et~al.(2026)Bhatia, Rezagholizadeh, Ashrafi~Fashi, Gilitschenski, Appia, and Barsoum]{prismbench}
Vansh Bhatia, Mehdi Rezagholizadeh, Parsa Ashrafi~Fashi, Igor Gilitschenski, Vikram Appia, and Emad Barsoum.
\newblock {PRISM-BENCH}: Diagnosing grounded and faithful multimodal reasoning beyond final-answer accuracy.
\newblock \emph{arXiv preprint arXiv:2610.36554}, 2026.

\bibitem[Cheng et~al.(2025)Cheng, Liu, Zhang, Fei, Hong, Lyu, Wang, Su, Gu, Liu, Bai, Tang, Wang, and Huang]{glyph}
Jiale Cheng, Yusen Liu, Xinyu Zhang, Yulin Fei, Wenyi Hong, Ruiliang Lyu, Weihan Wang, Zhe Su, Xiaotao Gu, Xiao Liu, Yushi Bai, Jie Tang, Hongning Wang, and Minlie Huang.
\newblock Glyph: Scaling context windows via visual-text compression.
\newblock \emph{arXiv preprint arXiv:2510.17800}, 2025.

\bibitem[Clavi{\'e} and Brand(2025)]{readbench}
Benjamin Clavi{\'e} and Florian Brand.
\newblock {ReadBench}: Measuring the dense text visual reading ability of vision-language models.
\newblock \emph{arXiv preprint arXiv:2505.19091}, 2025.

\bibitem[{DeepSeek-AI} et~al.(2025){DeepSeek-AI}, Guo, Yang, Zhang, et~al.]{deepseekr1}
{DeepSeek-AI}, Daya Guo, Dejian Yang, Haowei Zhang, et~al.
\newblock {DeepSeek-R1}: Incentivizing reasoning capability in {LLMs} via reinforcement learning.
\newblock \emph{arXiv preprint arXiv:2501.12948}, 2025.

\bibitem[Deng et~al.(2025)Deng, Yuan, Bu, Wang, Li, et~al.]{longdocurl}
Chao Deng, Jiale Yuan, Pi Bu, Peijie Wang, Zhong-Zhi Li, et~al.
\newblock {LongDocURL}: a comprehensive multimodal long document benchmark integrating understanding, reasoning, and locating.
\newblock In \emph{Proceedings of the 63rd Annual Meeting of the Association for Computational Linguistics (ACL)}, 2025.
\newblock arXiv:2412.18424.

\bibitem[Du et~al.(2025)Du, Tian, Ronanki, Rongali, Bodapati, et~al.]{ctxlen}
Yufeng Du, Minyang Tian, Srikanth Ronanki, Subendhu Rongali, Sravan Bodapati, et~al.
\newblock Context length alone hurts {LLM} performance despite perfect retrieval.
\newblock In \emph{Findings of the Association for Computational Linguistics: EMNLP 2025}, 2025.
\newblock arXiv:2510.05381.

\bibitem[Fan et~al.(2026)Fan, Zhou, Luo, Xu, Lu, Zhu, Chen, and Guo]{caps}
Cheng Fan, Junyi Zhou, Tingzhang Luo, RongJian Xu, Qiyanhui Lu, Mingjian Zhu, Hanting Chen, and Jianyuan Guo.
\newblock Reading is not reasoning: Bridging the agentic policy gap in vision-text compression.
\newblock \emph{arXiv preprint arXiv:2608.08960}, 2026.

\bibitem[Feng(2025)]{lookrecite}
Xisheng Feng.
\newblock Look, recite, then answer: Enhancing {VLM} performance via {Self-Generated} knowledge hints.
\newblock \emph{arXiv preprint arXiv:2512.00882}, 2025.

\bibitem[Fu et~al.(2024)Fu, Guo, Khalighinejad, Liu, Dhingra, Yogatama, Jia, and Neiswanger]{isobench}
Deqing Fu, Ruohao Guo, Ghazal Khalighinejad, Ollie Liu, Bhuwan Dhingra, Dani Yogatama, Robin Jia, and Willie Neiswanger.
\newblock {IsoBench}: Benchmarking multimodal foundation models on isomorphic representations.
\newblock In \emph{Conference on Language Modeling (COLM)}, 2024.
\newblock arXiv:2404.01266.

\bibitem[Gao et~al.(2026)Gao, Chen, Xu, Chen, Guan, and Zeng]{zerosenseeval}
Yonghan Gao, Zehong Chen, Lijian Xu, Jingzhi Chen, Jingwei Guan, and Xingyu Zeng.
\newblock Decoupling semantics from vision: A framework for faithful visual-text compression evaluation.
\newblock \emph{arXiv preprint arXiv:2608.01848}, 2026.

\bibitem[{Gemma Team} et~al.(2025){Gemma Team}, Kamath, Ferret, Pathak, et~al.]{gemma3}
{Gemma Team}, Aishwarya Kamath, Johan Ferret, Shreya Pathak, et~al.
\newblock Gemma 3 technical report.
\newblock \emph{arXiv preprint arXiv:2503.19786}, 2025.

\bibitem[Geva et~al.(2023)Geva, Bastings, Filippova, and Globerson]{geva}
Mor Geva, Jasmijn Bastings, Katja Filippova, and Amir Globerson.
\newblock Dissecting recall of factual associations in {Auto-Regressive} language models.
\newblock In \emph{Proceedings of the 2023 Conference on Empirical Methods in Natural Language Processing (EMNLP)}, 2023.
\newblock arXiv:2304.14767.

\bibitem[{GLM-V Team} et~al.(2025){GLM-V Team}, Hong, Yu, Gu, et~al.]{glm41v}
{GLM-V Team}, Wenyi Hong, Wenmeng Yu, Xiaotao Gu, et~al.
\newblock {GLM-4.5V} and {GLM-4.1V-Thinking}: Towards versatile multimodal reasoning with scalable reinforcement learning.
\newblock \emph{arXiv preprint arXiv:2507.01006}, 2025.

\bibitem[Hsieh et~al.(2024)Hsieh, Sun, Kriman, Acharya, Rekesh, et~al.]{ruler}
Cheng-Ping Hsieh, Simeng Sun, Samuel Kriman, Shantanu Acharya, Dima Rekesh, et~al.
\newblock {RULER}: What's the real context size of your {Long-Context} language models?
\newblock In \emph{Conference on Language Modeling (COLM)}, 2024.
\newblock arXiv:2404.06654.

\bibitem[Hu et~al.(2024)Hu, Xu, Ye, Yan, Zhang, et~al.]{docowl}
Anwen Hu, Haiyang Xu, Jiabo Ye, Ming Yan, Liang Zhang, et~al.
\newblock {mPLUG-DocOwl} 1.5: Unified structure learning for {OCR-free} document understanding.
\newblock In \emph{Findings of the Association for Computational Linguistics: EMNLP 2024}, 2024.
\newblock arXiv:2403.12895.

\bibitem[Hu et~al.(2026)Hu, Xu, and Hu]{lookzoom}
Zhensheng Hu, Guanglong Xu, and Guanqiang Hu.
\newblock The answer knows where to look: Zooming before {MLLMs} answer under visual-token budgets.
\newblock \emph{arXiv preprint arXiv:2610.32664}, 2026.

\bibitem[Jiang et~al.(2023)Jiang, Wu, Lin, Yang, and Qiu]{llmlingua}
Huiqiang Jiang, Qianhui Wu, Chin-Yew Lin, Yuqing Yang, and Lili Qiu.
\newblock {LLMLingua}: Compressing prompts for accelerated inference of large language models.
\newblock In \emph{Proceedings of the 2023 Conference on Empirical Methods in Natural Language Processing (EMNLP)}, 2023.
\newblock arXiv:2310.05736.

\bibitem[Jo et~al.(2026)Jo, Jang, and Mo]{loudclear}
Eunchan Jo, Seounghyun Jang, and Sangwoo Mo.
\newblock Loud and clear: Making visual evidence count in multimodal document {RAG}.
\newblock \emph{arXiv preprint arXiv:2610.21318}, 2026.

\bibitem[Kim et~al.(2022)Kim, Hong, Yim, Nam, Park, et~al.]{donut}
Geewook Kim, Teakgyu Hong, Moonbin Yim, Jeongyeon Nam, Jinyoung Park, et~al.
\newblock {OCR-free} document understanding transformer.
\newblock In \emph{European Conference on Computer Vision (ECCV)}, 2022.
\newblock arXiv:2111.15664.

\bibitem[Kim et~al.(2025)Kim, Kim, Kwon, Lee, Yun, and Song]{kvzip}
Jang-Hyun Kim, Jinuk Kim, Sangwoo Kwon, Jae~W. Lee, Sangdoo Yun, and Hyun~Oh Song.
\newblock {KVzip}: Query-agnostic {KV} cache compression with context reconstruction.
\newblock In \emph{Advances in Neural Information Processing Systems (NeurIPS)}, 2025.
\newblock arXiv:2505.23416.

\bibitem[Kwiatkowski et~al.(2019)Kwiatkowski, Palomaki, Redfield, Collins, Parikh, et~al.]{nq}
Tom Kwiatkowski, Jennimaria Palomaki, Olivia Redfield, Michael Collins, Ankur Parikh, et~al.
\newblock Natural questions: A benchmark for question answering research.
\newblock \emph{Transactions of the Association for Computational Linguistics}, 7:\penalty0 452--466, 2019.

\bibitem[Le-Duc et~al.(2026)Le-Duc, Bui, Ngo, Hoang, Song, Liu, Hoang, Wu, Ngo, Zhang, Diab, Xu, Savarese, and Wu]{bedbench}
Khai Le-Duc, Nguyen Kim~Hai Bui, An Ngo, Nam~D. Hoang, Yueqi Song, Jiarui Liu, Thai~Quoc Hoang, Fang Wu, Chris Ngo, Terry~Jingchen Zhang, Mona~T. Diab, Ran Xu, Silvio Savarese, and Xiao-Ming Wu.
\newblock Where do vision language models fail at {3D} reasoning?
\newblock \emph{arXiv preprint arXiv:2610.39639}, 2026.

\bibitem[Lee et~al.(2025)Lee, Yang, and Berg-Kirkpatrick]{opticalae}
Ivan~Yee Lee, Cheng Yang, and Taylor Berg-Kirkpatrick.
\newblock Optical context compression is just (bad) autoencoding.
\newblock \emph{arXiv preprint arXiv:2512.03643}, 2025.

\bibitem[Lee et~al.(2023)Lee, Joshi, Turc, Hu, Liu, et~al.]{pix2struct}
Kenton Lee, Mandar Joshi, Iulia Turc, Hexiang Hu, Fangyu Liu, et~al.
\newblock {Pix2Struct}: Screenshot parsing as pretraining for visual language understanding.
\newblock In \emph{International Conference on Machine Learning (ICML)}, 2023.
\newblock arXiv:2210.03347.

\bibitem[Li and Shi(2026)]{transcribe}
Bojie Li and Noah Shi.
\newblock Transcribe, then reason: Two-pass decomposition for multimodal review.
\newblock \emph{arXiv preprint arXiv:2609.18958}, 2026.

\bibitem[Li et~al.(2025{\natexlab{a}})Li, Xu, Zhao, and Peng]{dyfo}
Geng Li, Jinglin Xu, Yunzhen Zhao, and Yuxin Peng.
\newblock {DyFo}: A {Training-Free} dynamic focus visual search for enhancing {LMMs} in {Fine-Grained} visual understanding.
\newblock In \emph{IEEE/CVF Conference on Computer Vision and Pattern Recognition (CVPR)}, 2025{\natexlab{a}}.
\newblock arXiv:2504.14920.

\bibitem[Li et~al.(2026)Li, Xie, Yao, Wu, Hong, et~al.]{insightdoc}
Kaican Li, Weiyan Xie, Lewei Yao, Jiannan Wu, Lanqing Hong, et~al.
\newblock {InSight-doc}: Agentic visual perception for long-document understanding.
\newblock \emph{arXiv preprint arXiv:2608.10628}, 2026.

\bibitem[Li et~al.(2024)Li, Huang, Yang, Venkitesh, Locatelli, Ye, Cai, Lewis, and Chen]{snapkv}
Yuhong Li, Yingbing Huang, Bowen Yang, Bharat Venkitesh, Acyr Locatelli, Hanchen Ye, Tianle Cai, Patrick Lewis, and Deming Chen.
\newblock {SnapKV}: {LLM} knows what you are looking for before generation.
\newblock In \emph{Advances in Neural Information Processing Systems (NeurIPS)}, 2024.
\newblock arXiv:2404.14469.

\bibitem[Li et~al.(2025{\natexlab{b}})Li, Lan, and Zhou]{textorpixels}
Yanhong Li, Zixuan Lan, and Jiawei Zhou.
\newblock Text or pixels? it takes half: On the token efficiency of visual text inputs in multimodal {LLMs}.
\newblock \emph{arXiv preprint arXiv:2510.18279}, 2025{\natexlab{b}}.

\bibitem[Liang et~al.(2026{\natexlab{a}})Liang, Zheng, Wang, and Mao]{spiral}
Tianyu Liang, Xiangxi Zheng, Yilin Wang, and Dongxing Mao.
\newblock Same semantics, different paths: Self-improving alignment for vision-text compression.
\newblock In \emph{ACM International Conference on Multimedia (MM)}, 2026{\natexlab{a}}.
\newblock arXiv:2608.02109.

\bibitem[Liang et~al.(2026{\natexlab{b}})Liang, Ying, et~al.]{visualmerit}
Yunhao Liang, Ruixuan Ying, et~al.
\newblock Visual merit or linguistic crutch? a close look at {DeepSeek-OCR}.
\newblock \emph{arXiv preprint arXiv:2601.03714}, 2026{\natexlab{b}}.

\bibitem[Lin et~al.(2026)Lin, Luo, Wang, Zhu, Ma, Zhu, and Xiao]{readingright}
Dingyang Lin, Yingfeng Luo, Chenglong Wang, Chenwei Zhu, Anxiang Ma, Jingbo Zhu, and Tong Xiao.
\newblock Reading right, answering wrong: How visual configuration changes affect evidence use in {VLMs}.
\newblock \emph{arXiv preprint arXiv:2609.25770}, 2026.

\bibitem[Liu et~al.(2024)Liu, Lin, Hewitt, Paranjape, Bevilacqua, et~al.]{lostmiddle}
Nelson~F. Liu, Kevin Lin, John Hewitt, Ashwin Paranjape, Michele Bevilacqua, et~al.
\newblock Lost in the middle: How language models use long contexts.
\newblock \emph{Transactions of the Association for Computational Linguistics}, 12:\penalty0 157--173, 2024.
\newblock arXiv:2307.03172.

\bibitem[Liu et~al.(2026)Liu, Feng, Wang, Han, Cheng, et~al.]{vistabench}
Qing'an Liu, Juntong Feng, Yuhao Wang, Xinzhe Han, Yujie Cheng, et~al.
\newblock {VISTA-Bench}: Do {Vision-Language} models really understand visualized text as well as pure text?
\newblock \emph{arXiv preprint arXiv:2602.04802}, 2026.

\bibitem[Luan et~al.(2024)]{textcot}
Bozhi Luan et~al.
\newblock {TextCoT}: Zoom in for enhanced multimodal text-rich image understanding.
\newblock \emph{arXiv preprint arXiv:2404.09797}, 2024.

\bibitem[Lyu et~al.(2025)Lyu, Ma, and Chen]{pixelworld}
Zhiheng Lyu, Xueguang Ma, and Wenhu Chen.
\newblock {PixelWorld}: How far are we from perceiving everything as pixels?
\newblock \emph{arXiv preprint arXiv:2501.19339}, 2025.

\bibitem[Ma et~al.(2024)Ma, Zang, Chen, Chen, Jiao, et~al.]{mmlongbenchdoc}
Yubo Ma, Yuhang Zang, Liangyu Chen, Meiqi Chen, Yizhu Jiao, et~al.
\newblock {MMLongBench-Doc}: Benchmarking long-context document understanding with visualizations.
\newblock In \emph{Advances in Neural Information Processing Systems (NeurIPS), Datasets and Benchmarks Track}, 2024.
\newblock arXiv:2407.01523.

\bibitem[Nacson et~al.(2025)]{docvlm}
Mor~Shpigel Nacson et~al.
\newblock {DocVLM}: Make your {VLM} an efficient reader.
\newblock In \emph{Proceedings of the IEEE/CVF Conference on Computer Vision and Pattern Recognition (CVPR)}, 2025.

\bibitem[Neo et~al.(2025)Neo, Ong, Torr, Geva, Krueger, and Barez]{neo}
Clement Neo, Luke Ong, Philip Torr, Mor Geva, David Krueger, and Fazl Barez.
\newblock Towards interpreting visual information processing in {Vision-Language} models.
\newblock In \emph{International Conference on Learning Representations (ICLR)}, 2025.
\newblock arXiv:2410.07149.

\bibitem[Nikankin et~al.(2025)Nikankin, Arad, Gandelsman, and Belinkov]{nikankin}
Yaniv Nikankin, Dana Arad, Yossi Gandelsman, and Yonatan Belinkov.
\newblock Same task, different circuits: Disentangling modality-specific mechanisms in {VLMs}.
\newblock \emph{arXiv preprint arXiv:2506.09047}, 2025.

\bibitem[{nostalgebraist}(2020)]{logitlens}
{nostalgebraist}.
\newblock Interpreting {GPT}: the logit lens.
\newblock LessWrong blog post, 2020.

\bibitem[Pei et~al.(2026)Pei, Li, Guo, and Zhu]{vera}
Rongcan Pei, Huan Li, Fang Guo, and Qi Zhu.
\newblock {VERA}: Identifying and leveraging visual evidence retrieval heads in long-context understanding.
\newblock \emph{arXiv preprint arXiv:2602.10146}, 2026.

\bibitem[Peng et~al.(2026)Peng, Xia, Zhong, Zeng, Han, Zhou, Liu, Zhang, and Yao]{simpleocr}
Yibo Peng, Peng Xia, Ding Zhong, Kaide Zeng, Siwei Han, Yiyang Zhou, Jiaqi Liu, Ruiyi Zhang, and Huaxiu Yao.
\newblock {SimpleOCR}: Rendering visualized questions to teach {MLLMs} to read.
\newblock \emph{arXiv preprint arXiv:2602.22426}, 2026.

\bibitem[Robertson and Zaragoza(2009)]{bm25}
Stephen Robertson and Hugo Zaragoza.
\newblock The probabilistic relevance framework: {BM25} and beyond.
\newblock \emph{Foundations and Trends in Information Retrieval}, 3\penalty0 (4):\penalty0 333--389, 2009.

\bibitem[Rust et~al.(2023)Rust, Lotz, Bugliarello, Salesky, de~Lhoneux, and Elliott]{pixel}
Phillip Rust, Jonas~F. Lotz, Emanuele Bugliarello, Elizabeth Salesky, Miryam de Lhoneux, and Desmond Elliott.
\newblock Language modelling with pixels.
\newblock In \emph{International Conference on Learning Representations (ICLR)}, 2023.
\newblock arXiv:2207.06991.

\bibitem[Shahgir et~al.(2026)Shahgir, Chen, Fu, Shayegani, Abu-Ghazaleh, Kementchedjhieva, and Dong]{vlmsneedwords}
Haz~Sameen Shahgir, Xiaofu Chen, Yu Fu, Erfan Shayegani, Nael Abu-Ghazaleh, Yova Kementchedjhieva, and Yue Dong.
\newblock {VLMs} need words: Vision language models ignore visual detail in favor of semantic anchors.
\newblock In \emph{Conference on Language Modeling (COLM)}, 2026.
\newblock arXiv:2604.02486.

\bibitem[Shen et~al.(2025)Shen, Zhao, Zhao, Xu, Zhang, et~al.]{zoomeye}
Haozhan Shen, Kangjia Zhao, Tiancheng Zhao, Ruochen Xu, Zilun Zhang, et~al.
\newblock {ZoomEye}: Enhancing multimodal {LLMs} with {Human-Like} zooming capabilities through {Tree-Based} image exploration.
\newblock In \emph{Proceedings of the 2025 Conference on Empirical Methods in Natural Language Processing (EMNLP)}, 2025.
\newblock arXiv:2411.16044.

\bibitem[Sun et~al.(2026)Sun, Yuan, Liu, Zhao, Zhang, Dredze, and Bai]{readingnotthinking}
Kaiser Sun, Xiaochuang Yuan, Hongjun Liu, Chen Zhao, Cheng Zhang, Mark Dredze, and Fan Bai.
\newblock Reading, not thinking: Understanding and bridging the modality gap when text becomes pixels in multimodal {LLMs}.
\newblock \emph{arXiv preprint arXiv:2603.09095}, 2026.

\bibitem[Sun et~al.(2023)Sun, Wang, Tay, Yang, and Zhou]{recite}
Zhiqing Sun, Xuezhi Wang, Yi Tay, Yiming Yang, and Denny Zhou.
\newblock {Recitation-Augmented} language models.
\newblock In \emph{International Conference on Learning Representations (ICLR)}, 2023.
\newblock arXiv:2210.01296.

\bibitem[Trivedi et~al.(2022)Trivedi, Balasubramanian, Khot, and Sabharwal]{musique}
Harsh Trivedi, Niranjan Balasubramanian, Tushar Khot, and Ashish Sabharwal.
\newblock {MuSiQue}: Multihop questions via single-hop question composition.
\newblock \emph{Transactions of the Association for Computational Linguistics}, 10:\penalty0 539--554, 2022.
\newblock arXiv:2108.00573.

\bibitem[van Sprang et~al.(2026)van Sprang, Samson, Lucic, Acar, Ghebreab, and Asano]{rest}
Angela van Sprang, Laurens Samson, Ana Lucic, Erman Acar, Sennay Ghebreab, and Yuki~M. Asano.
\newblock Same content, different answers: Cross-modal inconsistency in {MLLMs}.
\newblock In \emph{Proceedings of the IEEE/CVF Conference on Computer Vision and Pattern Recognition (CVPR)}, 2026.
\newblock arXiv:2512.08923.

\bibitem[Veikherman et~al.(2026)Veikherman, Ventura, Calderon, Zelnik-Manor, and Reichart]{encodedunused}
Anat Veikherman, Mor Ventura, Nitay Calderon, Lihi Zelnik-Manor, and Roi Reichart.
\newblock Encoded but unused: Bounding the sources of {VLM} errors.
\newblock \emph{arXiv preprint arXiv:2610.29241}, 2026.

\bibitem[Wang et~al.(2024)Wang, Li, Lin, Li, Wang, and Shou]{visincontext}
Alex~Jinpeng Wang, Linjie Li, Yiqi Lin, Min Li, Lijuan Wang, and Mike~Zheng Shou.
\newblock Leveraging visual tokens for extended text contexts in multi-modal learning.
\newblock \emph{arXiv preprint arXiv:2406.02547}, 2024.

\bibitem[Wang et~al.(2025)Wang, Gao, Gu, Pu, Cui, et~al.]{internvl35}
Weiyun Wang, Zhangwei Gao, Lixin Gu, Hengjun Pu, Long Cui, et~al.
\newblock {InternVL3.5}: Advancing {Open-Source} multimodal models in versatility, reasoning, and efficiency.
\newblock \emph{arXiv preprint arXiv:2508.18265}, 2025.

\bibitem[Wei et~al.(2025)Wei, Sun, and Li]{deepseekocr}
Haoran Wei, Yaofeng Sun, and Yukun Li.
\newblock {DeepSeek-OCR}: Contexts optical compression.
\newblock \emph{arXiv preprint arXiv:2510.18234}, 2025.

\bibitem[Wei et~al.(2022)Wei, Wang, Schuurmans, Bosma, Ichter, et~al.]{cot}
Jason Wei, Xuezhi Wang, Dale Schuurmans, Maarten Bosma, Brian Ichter, et~al.
\newblock {Chain-of-Thought} prompting elicits reasoning in large language models.
\newblock In \emph{Advances in Neural Information Processing Systems (NeurIPS)}, 2022.
\newblock arXiv:2201.11903.

\bibitem[Wu and Xie(2024)]{vstar}
Penghao Wu and Saining Xie.
\newblock {V*}: Guided visual search as a core mechanism in multimodal {LLMs}.
\newblock In \emph{IEEE/CVF Conference on Computer Vision and Pattern Recognition (CVPR)}, 2024.
\newblock arXiv:2312.14135.

\bibitem[Xia et~al.(2026)Xia, Cheng, Bao, Du, and Liu]{foundnotread}
Qingtao Xia, Siyao Cheng, Jiahua Bao, Jiaxing Du, and Jie Liu.
\newblock Found but not read: When extracted text closes the retrieval-reading gap in document vision-language models.
\newblock \emph{arXiv preprint arXiv:2610.02880}, 2026.

\bibitem[Xie et~al.(2026)Xie, Friedman, Yu, Pan, Fifty, Kim, Du, Gan, Rathod, and Dhingra]{lensvlm}
Roy Xie, Dan Friedman, Donghan Yu, Bowen Pan, Christopher Fifty, Jang-Hyun Kim, Xianzhi Du, Zhe Gan, Vivek Rathod, and Bhuwan Dhingra.
\newblock {LensVLM}: Selective context expansion for compressed visual representation of text.
\newblock In \emph{Advances in Neural Information Processing Systems (NeurIPS)}, 2026.
\newblock arXiv:2605.07019.

\bibitem[Xing et~al.(2025)Xing, Wang, Yan, Shu, and Tang]{vist}
Ling Xing, Alex~Jinpeng Wang, Rui Yan, Xiangbo Shu, and Jinhui Tang.
\newblock Vision-centric token compression in large language model.
\newblock In \emph{Advances in Neural Information Processing Systems (NeurIPS)}, 2025.
\newblock arXiv:2502.00791.

\bibitem[Xu et~al.(2026{\natexlab{a}})Xu, Zhai, Fang, Xu, Lu, et~al.]{seer}
Jiawei Xu, Zhilin Zhai, Jinrui Fang, Ruohan Xu, Mingfei Lu, et~al.
\newblock {SEER}: Long-context reasoning via selective visual-text compression.
\newblock In \emph{Conference on Language Modeling (COLM)}, 2026{\natexlab{a}}.
\newblock arXiv:2608.15962.

\bibitem[Xu et~al.(2026{\natexlab{b}})Xu, Wang, Wu, Song, Lin, and Shen]{crossmath}
Yige Xu, Yongjie Wang, Zizhuo Wu, Kaisong Song, Jun Lin, and Zhiqi Shen.
\newblock Do vision-language models truly perform vision reasoning? {A} rigorous study of the modality gap.
\newblock \emph{arXiv preprint arXiv:2604.16256}, 2026{\natexlab{b}}.

\bibitem[Yang et~al.(2018)Yang, Qi, Zhang, Bengio, Cohen, et~al.]{hotpotqa}
Zhilin Yang, Peng Qi, Saizheng Zhang, Yoshua Bengio, William~W. Cohen, et~al.
\newblock {HotpotQA}: A dataset for diverse, explainable multi-hop question answering.
\newblock In \emph{Proceedings of the 2018 Conference on Empirical Methods in Natural Language Processing (EMNLP)}, 2018.
\newblock arXiv:1809.09600.

\bibitem[Zeng et~al.(2026)Zeng, Wang, Guo, Dai, Long, and Liu]{agar}
Shenglai Zeng, Qirui Wang, Kai Guo, Xinnan Dai, Xianxuan Long, and Hui Liu.
\newblock Magnifying what matters: Attention-guided adaptive rendering for visual text comprehension.
\newblock \emph{arXiv preprint arXiv:2606.12898}, 2026.

\bibitem[Zhang et~al.(2025{\natexlab{a}})Zhang, Khayatkhoei, Chhikara, and Ilievski]{vicrop}
Jiarui Zhang, Mahyar Khayatkhoei, Prateek Chhikara, and Filip Ilievski.
\newblock {MLLMs} know where to look: Training-free perception of small visual details with multimodal {LLMs}.
\newblock In \emph{International Conference on Learning Representations (ICLR)}, 2025{\natexlab{a}}.
\newblock arXiv:2502.17422.

\bibitem[Zhang et~al.(2025{\natexlab{b}})Zhang, Gao, Zhang, Li, Zhang, et~al.]{chainoffocus}
Xintong Zhang, Zhi Gao, Bofei Zhang, Pengxiang Li, Xiaowen Zhang, et~al.
\newblock Adaptive {Chain-of-Focus} reasoning via dynamic visual search and zooming for efficient {VLMs}.
\newblock \emph{arXiv preprint arXiv:2505.15436}, 2025{\natexlab{b}}.

\bibitem[Zhang et~al.(2025{\natexlab{c}})Zhang, Yadav, Han, and Shutova]{zhangflow}
Zhi Zhang, Srishti Yadav, Fengze Han, and Ekaterina Shutova.
\newblock Cross-modal information flow in multimodal large language models.
\newblock In \emph{IEEE/CVF Conference on Computer Vision and Pattern Recognition (CVPR)}, 2025{\natexlab{c}}.
\newblock arXiv:2411.18620.

\bibitem[Zhao et~al.(2025)Zhao, Wang, Zhu, Liu, Ni, et~al.]{vtcbench}
Hongbo Zhao, Meng Wang, Fei Zhu, Wenzhuo Liu, Bolin Ni, et~al.
\newblock {VTCBench}: Can {Vision-Language} models understand long context with {Vision-Text} compression?
\newblock \emph{arXiv preprint arXiv:2512.15649}, 2025.

\bibitem[Zheng et~al.(2026)Zheng, Yang, Hong, Zhao, Xu, et~al.]{deepeyes}
Ziwei Zheng, Michael Yang, Jack Hong, Chenxiao Zhao, Guohai Xu, et~al.
\newblock {DeepEyes}: Incentivizing ``thinking with images'' via reinforcement learning.
\newblock In \emph{International Conference on Learning Representations (ICLR)}, 2026.
\newblock arXiv:2505.14362.

\bibitem[Zhong et~al.(2026)Zhong, Qiu, Pan, Liu, Gu, et~al.]{focusvtc}
FangZhi Zhong, Xuerui Qiu, Yuqi Pan, Ya Liu, Shaowei Gu, et~al.
\newblock {FocusVTC}: Efficient and {High-Performance} visual text compression with adaptive resolution.
\newblock \emph{arXiv preprint arXiv:2609.36651}, 2026.

\end{thebibliography}
}

\clearpage
\onecolumn
\begin{center}
{\Large\textbf{\thetitle}}\\[0.5em]
{\large Supplementary Material}
\end{center}
\vspace{1em}
\appendix
\noindent The supplement has three parts. \textbf{Part A} (Methods) describes every experiment completely enough to reproduce it. \textbf{Part B} (Extended Data) gives the evidence behind each observation of the main text, one item per finding, each opening with its result. \textbf{Part C} holds the prompts and the release statement. \Cref{tab:roadmap} maps each observation to its methods and evidence.

\begin{table}[!ht]
\caption{\textbf{Where to find the evidence for each observation.}}
\label{tab:roadmap}
\centering\footnotesize
\setlength{\tabcolsep}{5pt}
\setlength{\aboverulesep}{0pt}\setlength{\belowrulesep}{0pt}
\renewcommand{\arraystretch}{1.12}
\begin{tabular}{p{0.62\linewidth}ll}
\toprule
\headrow \textbf{Observation} & \textbf{Methods} & \textbf{Evidence} \\
\midrule
\obs{1} VLMs recognize evidence they cannot use & \ref{app:m-gap} & \ref{app:readers}, \ref{app:alt}, \ref{app:mech} \\
\obs{2} Recognizing visual text needs fewer pixels than using it & \ref{app:m-gap} & \ref{app:heights} \\
\obs{3} Thinking closes the gap by putting the answer into words & \ref{app:m-think} & \ref{app:think}, \ref{app:cf} \\
\obs{4} The answer reads the image before its characters become words & \ref{app:m-mech} & \ref{app:mech} \\
\obs{5} The read--use gap ranks how much words beat pixels & \ref{app:m-pred} & \ref{app:prereg} \\
\obs{6}--\obs{7} \ours finds the evidence and leads at equal budget & \ref{app:impl}, \ref{app:m-eval} & \ref{app:abs}, \ref{app:ablation} \\
\obs{8}--\obs{9} Transfer, and where words stop helping & \ref{app:m-eval} & \ref{app:forms} \\
\obs{10} Cost & \ref{app:m-eval} & \ref{app:cost} \\
\bottomrule
\end{tabular}
\end{table}

\section{Methods}\label{app:methods}

\subsection{Readers and Rendered Evidence}\label{app:m-readers}
\textbf{Evidence.} For the analyses of \cref{sec:gap}, each question keeps only its gold supporting evidence from long-document QA in the LensVLM setting (HotpotQA, MuSiQue and NQ, 100 questions each). The evidence is rendered at the density of $10\times$ visual-text compression, with characters about 6\,px tall, and the same image is enlarged by bicubic interpolation to 9, 15 and 24\,px. Every size therefore carries the same information and differs only in visual tokens per character. We give resolution as character height throughout.

\textbf{Readers.} Instruction-tuned: Qwen3-VL 2B/4B/8B/32B, InternVL3.5-8B (InternViT encoder) and Gemma-3-12B (SigLIP encoder). Thinking: GLM-4.1V-9B-Thinking, Qwen3-VL-8B-Thinking, Glyph (trained from the GLM-4.1V base on rendered text) and Elorian Glimpse. Frontier: GPT-6.1 (gpt-6.1-sol) and Gemini~4 (gemini-4-argon) on a 90-question subset (30 per dataset, the same images), and Glimpse through OpenRouter, all accessed in October 2026. Open readers decode greedily; thinking readers think as usual, up to 4{,}096 tokens (Glimpse: 8{,}192 through its API). InternVL tiles every image into fixed 448\,px blocks and Gemma resizes it to a fixed input, so for them enlarging changes little of what they see.

\textbf{Three forms of the same evidence.} Every reader answers each image from the original evidence text (Text), from the image in one pass (Image), and from its own transcript of the image, given as text without the image (Transcript). Prompts are in \cref{app:prompts}.

\subsection{Recognition, the Gap and Statistics}\label{app:m-gap}
\textbf{Recognition.} A question is recognized by a reader at a character height when the gold answer appears verbatim in a second transcript the reader makes of the same image with another prompt, apart from the transcript it answers from. Judging recognition on an independent transcript keeps the answering transcript from selecting its own successes. On MMLongBench-Doc evidence pages, where many answers are computed or summarized rather than copied, recognition is judged by an LLM (DeepSeek~V4.1 Flash).

\textbf{The read--use gap} of reader $m$ at height $r$ is $G(m,r) = \mathrm{Acc}_{\text{transcript}} - \mathrm{Acc}_{\text{image}}$, both on the questions $m$ recognizes at $r$. For Glimpse, questions where both answers are non-empty.

\textbf{Metric and statistics.} Accuracy is substring exact match (SubEM) unless stated. Confidence intervals are 95\% bootstrap intervals over questions (at least 2{,}000 resamples); differences between two conditions are paired, resampling the same questions. Rank correlations are Spearman's $\rho$ with exact or permutation $p$-values, as stated with each.

\subsection{Thinking Interventions}\label{app:m-think}
\textbf{Ablation.} Same 15\,px images. On: thinking as usual. Off: an empty thinking block is prefilled and the reader answers at once (for GLM-4.1V the prefill continues into the answer box, since an empty thought alone lets it reason inside the answer). Prompts are rendered from the user turn with the model's chat template, and the prefill is appended as raw text. A trace \emph{quotes} the evidence when it shares a run of at least six consecutive words with the reader's own transcript.

\textbf{Trace edits.} We use traces whose thinking closed, on recognized questions. A word is \emph{quoted} if it lies in a run of at least six consecutive words shared with the reader's own transcript, and is an \emph{answer} word if it belongs to a mention of the answer the reader gave with normal thinking; the gold answer is not used. \emph{Quotes only} keeps the quoted runs in order, joined by ellipses; \emph{answer only} keeps only the answer words inside them (an empty thought when the quotes hold no answer); \emph{other quoted words} keeps as many quoted words as answer only, the first quoted words that are not answer words; \emph{quotes without answer} keeps the quoted runs minus their answer words. \emph{Reasoning} deletes the quoted runs, and \emph{without answer} also every answer word. The edited trace is prefilled as the thinking, and the reader answers greedily. All conditions are run together in one environment.

\textbf{Counterfactual prefill.} On the same recognized questions, a single entity is prefilled as the thinking: the gold answer, or a wrong entity from the same evidence, namely the span of the reader's own transcript of the same kind as the gold answer (a number for a number, a name for a name), closest to it in length, that contains neither the gold answer nor the reader's own answer (found for 192 and 160 questions).

\textbf{Candidates in words.} For instruction-tuned readers, the question states the answer among three other entities from the same evidence (spans of the gold evidence of the same kind as the answer, closest to it in length, in shuffled order; 286 of the 300 questions have three such entities), and the reader answers again from the image and from its own transcript at 15\,px. Candidates alone, with no evidence, already give the answer 63--78\% of the time, so results are also reported on the recognized questions they do not give away.

\subsection{Controls}\label{app:m-controls}
\textbf{Language prior.} Each Qwen reader is asked the question alone, without evidence, and recognized questions it answers correctly this way are removed ($10\times$ evidence enlarged $2.5\times$, the protocol of the mechanistic probes).

\textbf{Synthetic records.} Rendered at 15\,px: four people with a password, a city and a pet each, mixed with four neutral sentences, with common names and random values. Questions use the evidence's own words (``What is the password of Kate Clark?'') or a rewording (``Which secret word does Kate Clark type to unlock their account?''). Questions count when the reader transcribes both the name and the answer.

\textbf{Words versus tokens.} 60 questions per dataset ($n = 180$), Qwen3-VL-8B: answering directly; adding the reader's own transcript to the prompt; transcribing then answering in the same generation; prefilling its own transcript; prefilling an equally long run of dots; prefilling another question's transcript; and reasoning freely before answering.

\textbf{Layout and distraction.} Answers split across rendered lines (21 questions); giving each line its own row of visual tokens; blanking the evidence lines in the full page while also showing the enlarged evidence.

\subsection{Mechanistic Probes}\label{app:m-mech}
\textbf{Attention cuts.} We cut attention from every position after the evidence (question and answer) to the evidence, layer by layer, and measure the drop in the answer's log-probability, normalized to cutting all layers. Qwen3-VL readers, $10\times$ gold evidence as text or enlarged $2.5\times$, 100 questions per dataset, on questions answered correctly in that form where the evidence matters. The answer reads the evidence from the first layer where cutting removes at least 10\% of the full-cut effect to the last such layer.\looseness=-1

\textbf{Logit lens.} Each visual token's hidden state is projected through the output head at every layer; we record when the words drawn in its patch enter the top five, against words from elsewhere in the same image as control. \emph{Words form} at the layer where this rate reaches half its peak.

\textbf{In the failing pass.} Visual tokens precede the question, so what they encode does not depend on it. On recognized questions at 15\,px, we apply the logit lens to the visual tokens that draw the answer, in the same forward pass that answers from the image, separately for correct and wrong answers.

\textbf{Patching.} The visual tokens' states at layers 8--16 are replaced with their states from layers 16--28 (22 layer combinations, with and without norm matching, 57 questions), or the states after the characters become words (layers 30--34) are placed at layers 12--20 (254 recognized questions).

\subsection{Prediction Study and Registrations}\label{app:m-pred}
\textbf{Test bed.} MMLongBench-Doc whole documents, text-type evidence (576 questions). Every condition sees the whole document at 48\,DPI; the second look adds the two pages chosen by a shared scan, as pixels at 144 or 48\,DPI or as the reader's own transcript of those pixels (words). Answers are judged by DeepSeek~V4.1 Flash. $\Delta = $ words $-$ pixels.

\textbf{Registrations.} Predictions were written before the corresponding outcomes and all outcomes are reported in \cref{app:prereg}.
\emph{Original registration} (six readers): the 15\,px gap on the 76 questions all six readers recognize is a fixed reader property; words beat pixels when it exceeds $+3$, and $\Delta$ follows the gap's rank.
\emph{Addendum 1} (before any outcome): extend to nine readers (Glimpse was later dropped for API cost; none of its outcomes was seen); primary test: Spearman $\rho > 0$, one-sided $p < 0.05$.
\emph{Addendum 2} (after Qwen3-VL-8B's 144\,DPI outcome, before all others): take the gap at the matched character height (24\,px $\leftrightarrow$ 144\,DPI, 6\,px $\leftrightarrow$ 48\,DPI), add a 48\,DPI arm, exclude Qwen3-VL-8B's 144\,DPI point from the confirmatory test, and make no prediction where recognition is below 20\%.
The registered predictors judge recognition by the transcript the reader answers from, as before the independent second transcripts of \cref{app:m-gap}; analyses with the second transcripts are post hoc.

\subsection{\texorpdfstring{\ours}{ALOUD}}\label{app:impl}
\textbf{Readout.} For each row of visual tokens, the features are the attention that the question tokens pay to the row, summed over the row's tokens, one feature per layer and head of the reader. A row is positive if it holds part of the gold evidence, located by rendering; these labels are used only on calibration questions. The readout is a logistic regression on standardized features, with the positive class weighted by the ratio of negative to positive rows and an L2 penalty chosen from $\{0.01, 0.03, 0.1, 0.3\}$ by four-fold cross-validation split by question. One readout per compression level is fitted on the calibration questions of all three datasets together and used unchanged on every test set and on Glyph's benchmarks; the 4B reader's readout is fitted the same way. A page scores as its best row.

\textbf{Reading aloud.} The top pages (three at $5\times$ and $10\times$, four at $15\times$; \cref{tab:app-cal}) are enlarged to character heights of about 15\,px and transcribed one page per request with greedy decoding. For timing, each page is instead cut at the gaps between lines into strips of four lines, all transcribed in one batch; on the test questions this gives the same accuracy as whole pages (54.2, 39.3 and 30.3 against 54.8, 39.3 and 28.3; paired $-0.7$ $[-2.7, +1.3]$, $0.0$ $[-2.8, +2.8]$, $+2.0$ $[-1.3, +5.2]$) with about as many generated tokens.  %
 Each transcript is aligned to the page's rows and cut into passages of about 100 words; a passage's scan score is the best score among the rows it covers. Passages are ranked by reciprocal rank fusion of the scan score and BM25 against the question ($k = 60$) at $5\times$ and $10\times$, and by BM25 alone at $15\times$, then kept in this order while they fit the budget: the tokens of the compressed pages plus six enlarged lines, the budget BM25 over the original text also gets. The reader answers from the kept passages alone, in document order.

\textbf{Other settings.} \ours (1 page) reads aloud only the top page. Without the scan, for reference, the reader reads every page aloud, enlarged $2\times$, $2.5\times$ and $3\times$ at the three levels, and its passages are ranked by BM25 alone.

\textbf{Calibration.} The number of pages read and the passage ranking are chosen on 100 calibration questions per dataset, with scan scores out of fold (\cref{tab:app-cal}). The rule, fixed before the grid was run, takes the best mean accuracy and, within one point of it, the fewest pages. Reading beyond three pages adds little, since the budget is fixed and further pages bring distractors; ranking by the scan alone falls behind at $10\times$ and $15\times$, and fusing it with BM25 helps where the scan is still sharp.

\begin{table*}[!ht]
\caption{\textbf{Calibration of reading aloud: three pages suffice.} Mean accuracy (SubEM) over 100 calibration questions per dataset for each number of pages read and each ranking of the read passages: BM25 match to the question, the scan score, or both fused by reciprocal rank. Gen.: tokens generated per question in transcription. \textbf{Bold}: the configuration chosen by the rule fixed in advance.}
\label{tab:app-cal}
\centering\footnotesize
\setlength{\tabcolsep}{4.2pt}
\setlength{\aboverulesep}{0pt}\setlength{\belowrulesep}{0pt}
\renewcommand{\arraystretch}{1.12}
\begin{tabular}{crrrrrrrrrrrr}
\toprule
\headrow & \multicolumn{4}{c}{\textbf{5$\boldsymbol{\times}$}} & \multicolumn{4}{c}{\textbf{10$\boldsymbol{\times}$}} & \multicolumn{4}{c}{\textbf{15$\boldsymbol{\times}$}} \\
\cmidrule(lr){2-5}\cmidrule(lr){6-9}\cmidrule(lr){10-13}
\headrow \textbf{Pages} & \textbf{BM25} & \textbf{Scan} & \textbf{Fused} & \textbf{Gen.} & \textbf{BM25} & \textbf{Scan} & \textbf{Fused} & \textbf{Gen.} & \textbf{BM25} & \textbf{Scan} & \textbf{Fused} & \textbf{Gen.} \\
\midrule
1 & 39.7 & 39.7 & 39.7 & 440 & 32.7 & 32.3 & 32.3 & 587 & 25.7 & 25.7 & 25.7 & 497 \\  %
2 & 50.3 & 50.3 & 50.3 & 888 & 37.3 & 36.7 & 37.7 & 1{,}190 & 26.7 & 26.7 & 26.3 & 1{,}037 \\
3 & 53.3 & 53.7 & \textbf{54.0} & 1{,}337 & 42.0 & 37.3 & \textbf{43.3} & 1{,}784 & 30.0 & 27.7 & 29.0 & 1{,}573 \\
4 & 52.7 & 52.7 & 53.3 & 1{,}789 & 41.3 & 37.7 & 41.0 & 2{,}384 & \textbf{31.3} & 26.7 & 29.7 & 2{,}106 \\
5 & 52.7 & 51.3 & 54.0 & 2{,}232 & 42.3 & 38.0 & 42.7 & 2{,}994 & 31.0 & 26.7 & 29.3 & 2{,}614 \\
6 & 52.3 & 52.3 & 53.3 & 2{,}687 & 42.0 & 38.3 & 41.7 & 3{,}598 & 29.0 & 26.7 & 28.7 & 3{,}133 \\
8 & 51.3 & 52.7 & 54.7 & 3{,}583 & 42.0 & 38.3 & 42.0 & 4{,}771 & 30.7 & 26.7 & 29.0 & 4{,}196 \\
\bottomrule
\end{tabular}
\end{table*}

\textbf{Form of the evidence.} \ours hands over the found pages themselves when their native resolution already reaches the character height at which the reader's answers from the image catch up with reading aloud (24\,px for Qwen3-VL-8B, \cref{fig:density}a), as with scans at 144\,DPI, and reads them aloud otherwise. The height is fixed by rendering or scanning, before any question.

\subsection{Compressed-Document Evaluation}\label{app:m-eval}
\textbf{Setting.} The LensVLM evaluation: HotpotQA, MuSiQue and NQ documents of 3K--12K tokens rendered at its $5\times$, $10\times$ and $15\times$ presets; 200 test and 400 calibration questions per dataset; SubEM.

\textbf{Budget.} We budget what the reader answers from. The effective compression rate is the original-text tokens over the tokens of the context the answer is generated from, transcripts and thinking included (for LensVLM, its single growing context). Passes whose inputs are gone before the answer, such as our scan and reading aloud and ViCrop's reference pass, count as compute and are reported in \cref{app:cost}. Compute is not budgeted but is capped at the cost of reading the original text (8{,}847 tokens on average): a method whose discarded passes process or generate more saves nothing over reading the text, and is listed for reference only. The same-budget gain is accuracy minus that of uniform rendering at the same rate, interpolated linearly between nine densities (0.6--18$\times$); beyond the densest, the curve is held at its last value.

\textbf{Baselines without the original text} (ranked). Transcribe all: the same reader transcribes every page, enlarged as for the zoom agent below, with the prompt of reading aloud, or DeepSeek-OCR transcribes every compressed page with its official prompt, and the reader answers from the whole transcript. Thinking reader: Qwen3-VL-8B-Thinking on the same pages with its recommended sampling (temperature 0.6, top-$p$ 0.95, top-$k$ 20), up to 24{,}576 tokens. Zoom agent: the same reader, allowed to ask for up to two pages enlarged $2\times$, $2.5\times$ and $3\times$ at the three levels. DeepEyes: its official system prompt, tool parsing, cropping and stopping rules, with the pages tiled in reading order into a single image of eight pages per row, page numbers printed on each; its crops and earlier turns count toward the budget, as thinking does. ViCrop: attention of the answer position to the visual tokens, divided by the attention under a generic instruction, summed over windows of page width and six token rows; the best window is enlarged as above and appended, with its layer chosen on calibration questions. VERA with pixels and Glyph follow their official protocols.

\textbf{Methods that need the original text} (reference, not ranked). AGAR, VERA with text, LensVLM-9B, and BM25 over the original text cut into passages of about 100 words with the budget of \ours.

\textbf{Methods above the cap} (reference, not ranked). Every page transcribed, then searched with BM25 within the budget of \ours: by our reader, which is \ours without the scan, or by DeepSeek-OCR on every compressed page with its official prompt. They generate 9{,}389, 9{,}478 and 11{,}921 (our reader) and 9{,}896, 12{,}566 and 26{,}364 (DeepSeek-OCR) tokens per question at the three levels.

\textbf{Transfer.} In every transfer setting the reader keeps its own input and the evidence is placed beside it. For Glyph, the passages \ours keeps fill up to as many tokens as Glyph's own pages, counted with its tokenizer, on its own rendering of LongBench (DPI 72 and 48) and RULER. For the frontier readers, the passages of our 8B reader are given with the compressed pages. On MMLongBench-Doc, the two scanned pages at 144\,DPI are added to the whole document at 48\,DPI and compared with uniform 72\,DPI rendering.

\textbf{Timing.} One question at a time on one RTX 6000D with vLLM. For timing, each found page is cut at the gaps between lines into strips of four lines, all transcribed in one batch.

\section{Extended Data}\label{app:data}

\subsection{Every Instruction-Tuned Reader Has the Gap}\label{app:readers}
\begin{table}[!ht]
\caption{\textbf{Every instruction-tuned reader has a read--use gap; no thinking reader's gap is significantly positive.} Gold evidence at 15\,px character height, 300 questions, the same images for every reader. Recognized: questions whose second transcript of the image, made with another prompt, contains the gold answer; Transcript answers from the first. Text, Transcript and Image are accuracies on recognized questions; Gap $=$ Transcript $-$ Image, with 95\% CI. Thinking readers think as usual, greedy, up to 4096 tokens (Glimpse: 8192, through its API); GLM-4.1V and Qwen3-VL-8B-Thinking use the same run as \cref{tab:app-think}. Glimpse: questions where both answers are non-empty. \textbf{Bold}: CI above zero. All: Transcript $-$ Image over all questions, recognized or not (instruction-tuned readers: CI above zero for the 4B, 8B and 32B readers).}
\label{tab:app-readers}
\centering\footnotesize
\setlength{\tabcolsep}{5pt}
\setlength{\aboverulesep}{0pt}\setlength{\belowrulesep}{0pt}
\renewcommand{\arraystretch}{1.12}
\begin{tabular}{p{2.0cm}lccccrlr}
\toprule
\headrow \textbf{Group} & \textbf{Reader} & \textbf{Recognized (\%)} & \textbf{Text} & \textbf{Transcript} & \textbf{Image} & \multicolumn{1}{c}{\textbf{Gap}} & \textbf{95\% CI} & \textbf{All} \\
\midrule
 & Qwen3-VL-2B & 62 & 57.0 & 52.7 & 41.4 & \textbf{+11.3} & \cint{+3.2}{+19.4} & +4.0 \\  %
 & Qwen3-VL-4B & 70 & 64.6 & 60.3 & 46.4 & \textbf{+13.9} & \cint{+6.2}{+21.5} & +7.3 \\  %
 & Qwen3-VL-8B & 84 & 70.8 & 66.4 & 55.7 & \textbf{+10.7} & \cint{+4.7}{+16.6} & +7.7 \\  %
 & Qwen3-VL-32B & 88 & 74.6 & 73.5 & 59.5 & \textbf{+14.0} & \cint{+8.7}{+19.3} & +11.7 \\  %
 & InternVL3.5-8B & 65 & 69.9 & 68.4 & 56.1 & \textbf{+12.2} & \cint{+5.1}{+18.9} & +5.3 \\  %
\multirow{-6}{2.0cm}{\raggedright\itshape Instruction-tuned} & Gemma-3-12B & 80 & 68.0 & 66.0 & 58.9 & \textbf{+7.1} & \cint{+1.2}{+13.3} & +5.0 \\  %
\midrule
 & GLM-4.1V-9B-Thinking & 65 & 80.6 & 68.4 & 69.9 & $-$1.5 & \cint{$-$8.2}{+5.6} & $-$5.3 \\  %
 & Qwen3-VL-8B-Thinking & 66 & 74.9 & 69.8 & 63.8 & +6.0 & \cint{$-$0.5}{+13.1} & $-$1.0 \\  %
 & Glyph & 60 & 78.8 & 62.0 & 72.1 & $-$10.1 & \cint{$-$17.3}{$-$2.8} & $-$9.7 \\  %
\multirow{-4}{2.0cm}{\raggedright\itshape Thinking} & Glimpse & 79 & 81.2 & 76.9 & 80.8 & $-$3.8 & \cint{$-$8.7}{+1.0} & $-$5.3 \\  %
\bottomrule
\end{tabular}
\end{table}

Frontier readers have smaller gaps (\cref{tab:app-frontier}); Gemini~4, a thinking reader, has none, as \obs{3} explains. Glyph recites the document in its thinking whatever it is asked, so its answers from the image already pass through its own words; at 32B the gap is still $+5.8$ $[+1.4, +10.1]$ at 24\,px (277 recognized questions). %

\begin{table}[!ht]
\caption{\textbf{Frontier readers have small gaps}, 90 questions, 15\,px. Columns as in \cref{tab:app-readers}.}
\label{tab:app-frontier}
\centering\footnotesize
\setlength{\tabcolsep}{5pt}
\setlength{\aboverulesep}{0pt}\setlength{\belowrulesep}{0pt}
\renewcommand{\arraystretch}{1.12}
\begin{tabular}{lccccrlr}
\toprule
\headrow \textbf{Reader} & \textbf{Recognized} & \textbf{Text} & \textbf{Transcript} & \textbf{Image} & \multicolumn{1}{c}{\textbf{Gap}} & \textbf{95\% CI} & \textbf{All} \\
\midrule
GPT-6.1 & 93\% (84) & 81.0 & 76.2 & 72.6 & +3.6 & \cint{$-$2.4}{+10.7} & +4.4 \\  %
Gemini~4 & 98\% (88) & 85.2 & 81.8 & 84.1 & $-$2.3 & \cint{$-$6.8}{+2.3} & $-$2.2 \\  %
\bottomrule
\end{tabular}
\end{table}

\subsection{Recognition and Use Across Character Heights}\label{app:heights}
\begin{figure}[!ht]
\centering
\includegraphics[width=\linewidth]{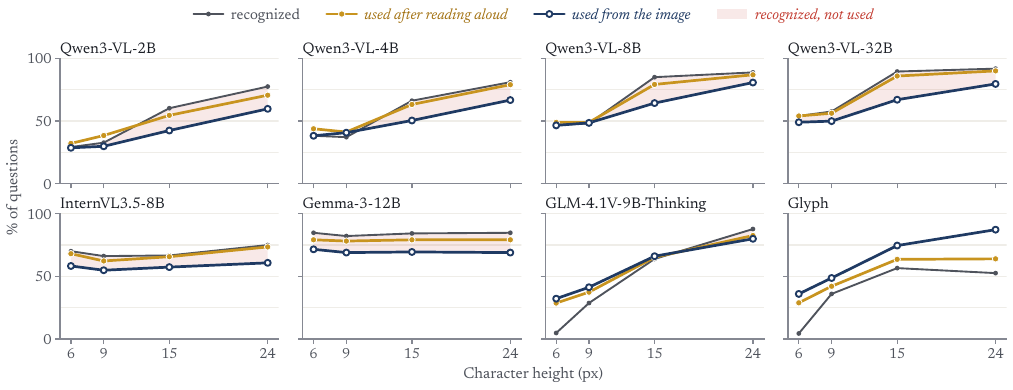}
\caption{\textbf{Recognition and use across character heights for eight readers} (text-answerable questions). At 6--9\,px, the recognition of GLM-4.1V and Glyph is a lower bound: their transcripts there often end inside unfinished thinking. InternVL tiles every image into fixed 448\,px blocks and Gemma resizes it to a fixed input, so enlarging changes little of what they see.}
\label{fig:grid}
\end{figure}

Across readers, recognition saturates before use (\cref{fig:grid}). On real document pages the gap appears as resolution falls and is largest where the evidence is text; on purely pictorial evidence, with nothing to recite, transcribing hurts (\cref{tab:app-docpages}).

\begin{table}[!ht]
\caption{\textbf{On real documents the gap grows as resolution falls}: the gap of Qwen3-VL-8B on MMLongBench-Doc evidence pages (1--3 pages per question) by evidence source, on recognized questions ($n$ at 48\,DPI). \textbf{Bold}: 95\% CI above zero.}
\label{tab:app-docpages}
\centering\footnotesize
\setlength{\tabcolsep}{6pt}
\setlength{\aboverulesep}{0pt}\setlength{\belowrulesep}{0pt}
\renewcommand{\arraystretch}{1.12}
\begin{tabular}{lrrrr}
\toprule
\headrow \textbf{Evidence} & \textbf{144\,DPI} & \textbf{72\,DPI} & \textbf{48\,DPI} & $\boldsymbol{n}$ \\
\midrule
Text & 0.0 & +7.2 & \textbf{+25.8} & 93 \\  %
Layout & $-$13.9 & 0.0 & +9.4 & 32 \\
Table & $-$3.0 & +4.9 & \textbf{+12.0} & 117 \\  %
Chart & $-$10.9 & +4.0 & \textbf{+16.1} & 93 \\  %
Picture & $-$14.8 & $-$5.4 & $-$8.3 & 108 \\
\midrule
All & $-$7.6 & +2.5 & \textbf{+10.6} & \\  %
\bottomrule
\end{tabular}
\end{table}

\subsection{The Gap Under Controls}\label{app:alt}
\label{app:prior}
Removing questions answerable from the question alone widens the gap at every scale (\cref{tab:app-prior}): such questions can be guessed from the image too, which masks it. \Cref{tab:app-controls} collects the other controls.

\begin{table}[!ht]
\caption{\textbf{Read--use gap before and after removing questions answerable from the question alone.} Qwen readers on $10\times$ evidence enlarged $2.5\times$, the protocol of the mechanistic probes. Prior only: accuracy from the question alone, over all questions. Gap $=$ Transcript $-$ Image on recognized questions ($n$), with 95\% CI. \textbf{Bold}: the gap after removal.}
\label{tab:app-prior}
\centering\footnotesize
\setlength{\tabcolsep}{5pt}
\setlength{\aboverulesep}{0pt}\setlength{\belowrulesep}{0pt}
\renewcommand{\arraystretch}{1.12}
\begin{tabular}{lcc>{\hspace{4pt}}crl>{\hspace{4pt}}crl}
\toprule
\headrow & & & \multicolumn{3}{c}{\textbf{Recognized}} & \multicolumn{3}{c}{\textbf{Recognized, prior wrong}} \\
\cmidrule(lr){4-6}\cmidrule(lr){7-9}
\headrow \multirow{-2}{*}{\textbf{Reader}} & \multirow{-2}{*}{\textbf{Questions}} & \multirow{-2}{*}{\textbf{Prior only}} & \textbf{$\boldsymbol{n}$} & \multicolumn{1}{c}{\textbf{Gap}} & \textbf{95\% CI} & \textbf{$\boldsymbol{n}$} & \multicolumn{1}{c}{\textbf{Gap}} & \textbf{95\% CI} \\
\midrule
Qwen3-VL-2B & 300 & 12.7 & 186 & +12.4 & \cint{+4.3}{+20.4} & 155 & \textbf{+16.1} & \cint{+7.1}{+24.5} \\  %
Qwen3-VL-4B & 300 & 17.7 & 209 & +14.8 & \cint{+7.7}{+22.0} & 166 & \textbf{+16.9} & \cint{+8.4}{+25.3} \\  %
Qwen3-VL-8B & 280 & 21.1 & 236 & +9.3 & \cint{+3.8}{+14.8} & 181 & \textbf{+11.0} & \cint{+4.4}{+17.7} \\  %
Qwen3-VL-32B & 150 & 22.7 & 134 & +10.4 & \cint{+3.7}{+17.9} & 101 & \textbf{+15.8} & \cint{+6.9}{+24.8} \\  %
\bottomrule
\end{tabular}
\end{table}

\begin{table}[!ht]
\caption{\textbf{Controls leave the gap in place; only more pixels narrow it.} Qwen3-VL-8B at 15\,px unless stated; protocols in \cref{app:m-controls}. Brackets: 95\% CIs.}
\label{tab:app-controls}
\centering\footnotesize
\setlength{\tabcolsep}{4pt}
\setlength{\aboverulesep}{0pt}\setlength{\belowrulesep}{0pt}
\renewcommand{\arraystretch}{1.12}
\begin{tabular}{p{0.44\linewidth}p{0.48\linewidth}}
\toprule
\headrow \textbf{Control} & \textbf{Result} \\
\midrule
Synthetic records, no prior knowledge, asked in the evidence's words & gap $+15.4$ $[+9.9, +21.4]$; image 83.0, own transcript 98.4 (182 of 210 questions) \\
\quad the same, question reworded & change in the gap $+0.0$ $[-3.8, +3.8]$ \\
\midrule
Reasoning freely before answering & $+3.0$ on recognized questions (CI crosses zero) \\
Transcribing first & $+7.6$ $[+2.1, +13.6]$ \\
Prefilling own transcript vs.\ as many dots & $+6.1$ $[+0.6, +12.2]$ \\
Prefilling own transcript vs.\ another question's & $+12.2$ $[+5.6, +18.9]$ \\
\midrule
Strict exact match / token F1 & transcribing first 55.1 / 69.3, image 47.4 / 61.3 \\
Answers split across lines (21 questions) & pixels 61.9, text 61.9 \\
One row of visual tokens per line (297 tokens) & $+5.6$, less than enlarging $5\times$ (572 tokens, $+8.7$) \\
Evidence lines blanked on the page & 43.6 vs.\ 43.6 \\
Enlarging the evidence $5\times$ (572 tokens) & gap $+3.7$ $[-0.9, +8.7]$; reading aloud closes it with about 200 text tokens \\
Depth doubled, 36 to 64 layers (8B to 32B) & gap not narrowed (\cref{tab:app-readers}) \\
\bottomrule
\end{tabular}
\end{table}

\textbf{What is said matters, not how much} ($n = 180$): answering directly gives 53.9; adding the reader's own transcript to the prompt 61.1 ($+7.2$); transcribing then answering in the same generation 58.9 ($+5.0$); prefilling its own transcript 57.8; prefilling an equally long run of dots 51.7; prefilling another question's transcript 45.6 ($-8.3$). Answers are short phrases (2.7--2.8 words on average), and transcribing then answering within one generation also helps, so a shared output budget~\citep{transcribe} does not apply. Wrong answers from the image are mostly other entities from the same evidence (Brazil for Germany, a character's actress for the character); 71\% of them appear verbatim in the transcript. On synthetic records they are the value of another person or attribute (52\%) or a near-miss spelling of the right one (29\%).

\subsection{Thinking Closes the Gap Through the Answer in Words}\label{app:think}
Not letting a thinking reader speak first brings the gap back to the level of instruction-tuned readers (\cref{tab:app-think}). Cutting its own trace shows that the benefit travels with the answer in words (\cref{tab:app-edits,tab:app-paired}): whatever keeps the answer recovers most of the benefit, and whatever removes it, including as many quoted words that are not the answer, recovers none.

\begin{table}[!ht]
\caption{\textbf{Not letting a thinking reader speak first brings the gap back.} Same 15\,px images; accuracies on recognized questions (GLM-4.1V: 196; Qwen3-VL-8B-Thinking: 199). On: thinking as usual. Off: an empty thought is prefilled and the reader answers at once. Gap $=$ Transcript $-$ Image, with 95\% CI. Quotes: traces that share a run of at least six consecutive words with the reader's own transcript; Answer: traces that contain the gold answer (both over all 300 questions).}
\label{tab:app-think}
\centering\footnotesize
\setlength{\tabcolsep}{3pt}
\setlength{\aboverulesep}{0pt}\setlength{\belowrulesep}{0pt}
\renewcommand{\arraystretch}{1.12}
\begin{tabular}{lcccl>{\hspace{3pt}}cccl>{\hspace{3pt}}cc}
\toprule
\headrow & \multicolumn{4}{c}{\textbf{Thinking on}} & \multicolumn{4}{c}{\textbf{Thinking off}} & \multicolumn{2}{c}{\textbf{In the trace}} \\
\cmidrule(lr){2-5}\cmidrule(lr){6-9}\cmidrule(lr){10-11}
\headrow \multirow{-2}{*}{\textbf{Reader}} & \textbf{Image} & \textbf{Transcript} & \textbf{Gap} & \textbf{95\% CI} & \textbf{Image} & \textbf{Transcript} & \textbf{Gap} & \textbf{95\% CI} & \textbf{Quotes} & \textbf{Answer} \\
\midrule
GLM-4.1V-9B-Thinking & 69.9 & 68.4 & $-$1.5 & \cint{$-$8.2}{+5.6} & 51.5 & 62.8 & \textbf{+11.2} & \cint{+4.1}{+18.9} & 87\% & 70\% \\  %
Qwen3-VL-8B-Thinking & 63.8 & 69.8 & +6.0 & \cint{$-$0.5}{+13.1} & 46.2 & 59.3 & \textbf{+13.1} & \cint{+5.5}{+20.6} & 88\% & 68\% \\  %
\bottomrule
\end{tabular}
\end{table}

\begin{table}[!ht]
\caption{\textbf{The answer in words carries the benefit of thinking.} Accuracy on recognized questions with a closed trace (GLM-4.1V: 193; Qwen3-VL-8B-Thinking: 163) when the reader's own trace is edited and prefilled as its thinking; answer words are located by the reader's own answer, not the gold one. Rec.: share of the full trace's benefit recovered, (edit $-$ off) / (full $-$ off). Words: share of the trace's words kept (GLM-4.1V / Qwen3-VL-8B-Thinking). Own transcript: answering from the reader's own transcript, written without the question, with empty thinking.}
\label{tab:app-edits}
\centering\footnotesize
\setlength{\tabcolsep}{5pt}
\setlength{\aboverulesep}{0pt}\setlength{\belowrulesep}{0pt}
\renewcommand{\arraystretch}{1.12}
\begin{tabular}{lrrrrr}
\toprule
\headrow & & \multicolumn{2}{c}{\textbf{GLM-4.1V-9B-Thinking}} & \multicolumn{2}{c}{\textbf{Qwen3-VL-8B-Thinking}} \\
\cmidrule(lr){3-4}\cmidrule(lr){5-6}
\headrow \textbf{Prefilled thinking} & \textbf{Words} & \textbf{Acc.} & \textbf{Rec.} & \textbf{Acc.} & \textbf{Rec.} \\
\midrule
Off (empty) & 0 & 51.8 & 0\% & 51.5 & 0\% \\  %
Full trace & 100\% & 67.9 & 100\% & 67.5 & 100\% \\  %
\multicolumn{6}{l}{\textit{Answer kept}} \\
Quotes only & 26\,/\,30\% & 61.1 & 58\% & 62.6 & 69\% \\  %
Reasoning with answer & 75\,/\,70\% & 67.4 & 97\% & 65.6 & 88\% \\  %
Answer only & 2\,/\,3\% & 62.7 & 68\% & 64.4 & 81\% \\  %
\multicolumn{6}{l}{\textit{Answer cut}} \\
Quotes without answer & 24\,/\,27\% & 52.8 & 6\% & 50.3 & $-$8\% \\  %
Reasoning without answer & 69\,/\,66\% & 47.7 & $-$26\% & 52.8 & 8\% \\  %
Other quoted words & 2\,/\,3\% & 48.2 & $-$23\% & 49.7 & $-$12\% \\  %
\midrule
\oursrow Own transcript & -- & 63.2 & 71\% & 62.0 & 65\% \\  %
\bottomrule
\end{tabular}
\end{table}

\begin{table}[!ht]
\caption{\textbf{Paired differences behind \cref{tab:app-edits}.} Accuracy differences in points with 95\% CIs. The reasoning helps only through the answer it repeats.}
\label{tab:app-paired}
\centering\footnotesize
\setlength{\tabcolsep}{4pt}
\setlength{\aboverulesep}{0pt}\setlength{\belowrulesep}{0pt}
\renewcommand{\arraystretch}{1.12}
\begin{tabular}{lll}
\toprule
\headrow \textbf{Comparison} & \textbf{GLM-4.1V-9B-Thinking} & \textbf{Qwen3-VL-8B-Thinking} \\
\midrule
Own transcript $-$ full trace & $-4.7$ $[-11.4, +2.1]$ & $-5.5$ $[-12.3, +1.2]$ \\
Answer only $-$ other quoted words & $+14.5$ $[+9.3, +20.2]$ & $+14.7$ $[+8.6, +21.5]$ \\
Answer only $-$ full trace & $-5.2$ $[-10.4, -0.5]$ & $-3.1$ $[-8.0, +1.8]$ \\
Answer only $-$ quotes only & $+1.6$ $[-3.1, +6.2]$ & $+1.8$ $[-1.2, +4.9]$ \\
Quotes without answer $-$ off & $+1.0$ $[-4.1, +6.2]$ & $-1.2$ $[-6.7, +4.3]$ \\
Reasoning without answer $-$ off & $-4.1$ $[-9.8, +1.6]$ & $+1.2$ $[-4.9, +7.4]$ \\
Reasoning with $-$ without answer & $+19.7$ $[+13.0, +26.4]$ & $+12.9$ $[+6.1, +19.6]$ \\
\bottomrule
\end{tabular}
\end{table}

\subsection{Readers Recognize the Answer Once It Is in Words}\label{app:cf}
Prefilled as its thinking, the correct answer is taken up nearly always and a wrong entity from the same evidence rarely, even on questions the reader's own thinking got wrong (\cref{tab:app-cf}): the reader checks the words against the image, and recognizes the answer once it is said, yet does not say it on its own. Instruction-tuned readers fail partly the same way: stating the answer among candidates halves their gap, and the rest closes only with the evidence itself in words (\cref{tab:app-cand}).

\begin{table}[!ht]
\caption{\textbf{Thinking readers take up the correct answer when it is said, far more often than a wrong one.} A single entity prefilled as the thinking, on recognized questions; brackets: 95\% CIs.}
\label{tab:app-cf}
\centering\footnotesize
\setlength{\tabcolsep}{4pt}
\setlength{\aboverulesep}{0pt}\setlength{\belowrulesep}{0pt}
\renewcommand{\arraystretch}{1.12}
\begin{tabular}{lll}
\toprule
\headrow \textbf{Prefilled thinking} & \textbf{GLM-4.1V-9B-Thinking} & \textbf{Qwen3-VL-8B-Thinking} \\
\midrule
Gold answer: answer given & 90.2\% & 93.9\% \\
Wrong entity from the evidence: entity given & 23\% $[17, 29]$ & 37\% $[29, 44]$ \\
\quad accuracy, empty $\to$ wrong entity & 52.1 $\to$ 39.6 & 52.5 $\to$ 35.0 \\
\midrule
\multicolumn{3}{l}{\textit{Questions its own full thinking got wrong} (62 and 53)} \\
Gold answer: answer given & 69.4\% & 81.1\% \\
Empty thought: gold answer given & 17.7\% & 18.9\% \\
\quad paired difference & $+51.6$ $[+38.7, +62.9]$ & $+62.3$ $[+49.1, +75.5]$ \\
Gold minus wrong entity, same questions & $+48.4$ $[+35.5, +61.3]$ & $+36.5$ $[+21.2, +51.9]$ \\
\bottomrule
\end{tabular}
\end{table}

\begin{table}[!ht]
\caption{\textbf{Stating the answer among candidates halves the gap.} Read--use gap (Transcript $-$ Image) on recognized questions at 15\,px, without and with four candidates in the question (the answer and three entities of the same kind from the evidence), with 95\% CIs. Left: all recognized questions with candidates; right: those the candidates alone, without evidence, do not answer. Mean: unweighted over readers, CI by bootstrapping within each reader.}
\label{tab:app-cand}
\centering\footnotesize
\setlength{\tabcolsep}{3pt}
\setlength{\aboverulesep}{0pt}\setlength{\belowrulesep}{0pt}
\renewcommand{\arraystretch}{1.12}
\begin{tabular}{lrrrrrr}
\toprule
\headrow & \multicolumn{3}{c}{\textbf{All recognized}} & \multicolumn{3}{c}{\textbf{Not given away by candidates}} \\
\cmidrule(lr){2-4}\cmidrule(lr){5-7}
\headrow \textbf{Reader} & \textbf{$\boldsymbol{n}$} & \textbf{Gap} & \textbf{With cand.} & \textbf{$\boldsymbol{n}$} & \textbf{Gap} & \textbf{With cand.} \\
\midrule
Qwen3-VL-2B & 174 & $+9.2$ & $+4.6$ & 65 & $+3.1$ & $+9.2$ \\  %
Qwen3-VL-4B & 199 & $+17.1$ & $+2.0$ & 48 & $+20.8$ & $+0.0$ \\  %
Qwen3-VL-8B & 240 & $+9.6$ & $+2.9$ & 54 & $+33.3$ & $+18.5$ \\  %
InternVL3.5-8B & 183 & $+12.6$ & $+4.9$ & 40 & $+17.5$ & $+10.0$ \\  %
Gemma-3-12B & 228 & $+5.7$ & $+0.0$ & 58 & $+8.6$ & $+5.2$ \\  %
\midrule
Mean & & $+10.8$ & $+2.9$ & & $+16.7$ & $+8.6$ \\  %
\quad 95\% CI & & \cint{+7.7}{+14.2} & \cint{+0.7}{+5.1} & & \cint{+10.7}{+23.0} & \cint{+3.0}{+14.6} \\
\bottomrule
\end{tabular}
\end{table}

\subsection{The Answer Reads the Evidence Before Its Characters Are Words}\label{app:mech}
\textbf{Same layers for text and pixels.} In Qwen3-VL-8B, the answer reads text from layer 12 to 28 (medians; interquartile 24--30) and enlarged pixels from 14 to 30 (26--32). Cutting attention after layers 24, 30 and 32 removes 61\%, 14\% and 6\% of the effect for text, and 60\%, 19\% and 12\% for pixels. The window opens and closes at the same layers for both forms at every scale and closes 6--8 layers before the output (\cref{tab:app-mech}).

\textbf{Most of what the answer takes is taken before the characters become words.} The words drawn in a patch enter the top five only from layer 24 on, reach half their peak at layer 30 and peak at layers 32--34 (control words: about 2\%). Text enters as words. About 81\% of what the answer position takes from the image is taken before layer 30; for Qwen3-VL-32B, whose characters become words at layer 60, about 99\%. Enlarging the image changes how much becomes readable (peak 15\%, 27\%, 20\% at $1\times$, $2.5\times$, $5\times$), not when. In the answer position, the first answer token emerges at layer 31 from text and from the own transcript, 32 from enlarged pixels and 35 from raw pixels.

\textbf{The answer is recognized in the pass that fails to use it.} On the 253 recognized questions at 15\,px, the answer's words reach the top five on as many of its visual tokens when the answer is wrong as when it is right (peak 17\% in both; 4\% for other words of the same image on the same tokens), with the same layer profile, rising from layer 24 and highest at layers 32--34. What differs is the answer position: the gold answer enters its top five at some layer for 98\% of correct answers and 68\% of wrong ones.

\textbf{Patching late visual states does not help.} No layer combination improves accuracy: the best, chosen on the test questions, is $-1.8$ and the worst $-21$; placing the states after the characters become words at layers 12--20 gives $-13.8$ to $-32.7$. On general image tasks the same intervention recovers about a third of the gap~\citep{nikankin}.

\begin{table}[!ht]
\caption{\textbf{Layers in which the answer position reads the evidence, text versus pixels, across scale.} $10\times$ gold evidence, as text or enlarged $2.5\times$. Start$\to$end: median layers at which the answer position starts and stops reading the evidence, by the attention cuts of \cref{app:m-mech} (cutting before the start, or from the end on, removes less than 10\% of the full-cut effect). Words: layer at which the words drawn in the image reach half their peak top-5 rate under the logit lens. Read before words: share of the pixel-evidence effect taken before that layer (2B and 4B: interpolated between the neighbouring cut layers).}
\label{tab:app-mech}
\centering\footnotesize
\setlength{\tabcolsep}{6pt}
\setlength{\aboverulesep}{0pt}\setlength{\belowrulesep}{0pt}
\renewcommand{\arraystretch}{1.12}
\begin{tabular}{lccccc}
\toprule
\headrow & & \multicolumn{2}{c}{\textbf{Start$\boldsymbol{\to}$end}} & & \\
\cmidrule(lr){3-4}
\headrow \multirow{-2}{*}{\textbf{Reader}} & \multirow{-2}{*}{\textbf{Layers}} & \textbf{Text} & \textbf{Pixels} & \multirow{-2}{*}{\textbf{Words}} & \multirow{-2}{*}{\textbf{Read before words}} \\
\midrule
Qwen3-VL-2B & 28 & 10$\to$22 & 10$\to$22 & 23 & 79\% \\  %
Qwen3-VL-4B & 36 & 14$\to$30 & 12$\to$30 & 27 & 61\% \\  %
Qwen3-VL-8B & 36 & 12$\to$28 & 14$\to$30 & 30 & 81\% \\  %
Qwen3-VL-32B & 64 & 40$\to$56 & 40$\to$56 & 60 & 99\% \\  %
\bottomrule
\end{tabular}
\end{table}

\subsection{The Gap Ranks the Benefit of Words}\label{app:prereg}
\Cref{tab:app-prereg} gives every registered outcome. The registered ranking tests pass and the registered sign rules do not: the gap ranks how much words beat pixels, not whether they do. \Cref{tab:app-rank} shows that the ranking holds within each family and within each density, without conditioning on recognition, and with recognition judged independently, and that recognition rates do not rank the benefit.

\begin{table*}[!ht]
\caption{\textbf{All registered outcomes: the gap ranks the benefit of words.} Out-of-sample test on MMLongBench-Doc, text-type evidence (576 questions). Every condition sees the whole document at 48\,DPI; the second look adds the two pages chosen by a shared scan, as pixels at the stated DPI or as the reader's own transcript of those pixels (Words). $\Delta =$ Words $-$ Pixels, with 95\% CI. Predictors, all measured on long-document gold evidence: the 15\,px gap on the 76 questions the original six readers all recognize (original registration and Addendum 1), and the gap at the character height matched to the second look (Addendum 2). --: recognition below 20\% at 6\,px, no prediction (GLM-4.1V 4\%, Glyph 5\%); Glyph was not run at 48\,DPI. $^\dagger$Outcome seen before Addendum 2, excluded from its confirmatory test. \textbf{Bold}: CI above zero.}
\label{tab:app-prereg}
\centering\footnotesize
\setlength{\tabcolsep}{3.2pt}
\setlength{\aboverulesep}{0pt}\setlength{\belowrulesep}{0pt}
\renewcommand{\arraystretch}{1.12}
\begin{tabular}{lr>{\hspace{5pt}}rccrl>{\hspace{5pt}}rccrl}
\toprule
\headrow & & \multicolumn{5}{c}{\textbf{Second look at 144\,DPI}} & \multicolumn{5}{c}{\textbf{Second look at 48\,DPI}} \\
\cmidrule(lr){3-7}\cmidrule(lr){8-12}
\headrow \multirow{-2}{*}{\textbf{Reader}} & \multicolumn{1}{c}{\multirow{-2}{*}{\textbf{15\,px gap}}} & \multicolumn{1}{c}{\hspace{5pt}\textbf{24\,px gap}} & \textbf{Pixels} & \textbf{Words} & \multicolumn{1}{c}{$\boldsymbol{\Delta}$} & \textbf{95\% CI} & \multicolumn{1}{c}{\hspace{5pt}\textbf{6\,px gap}} & \textbf{Pixels} & \textbf{Words} & \multicolumn{1}{c}{$\boldsymbol{\Delta}$} & \textbf{95\% CI} \\
\midrule
Qwen3-VL-2B & +5.3 & +8.3 & 31.8 & 28.3 & $-$3.5 & \cint{$-$6.4}{$-$0.3} & +20.0 & 17.0 & 19.6 & +2.6 & \cint{+0.0}{+5.4} \\  %
Qwen3-VL-4B & +7.9 & +10.2 & 41.3 & 39.4 & $-$1.9 & \cint{$-$4.7}{+0.9} & +24.3 & 20.1 & 25.9 & \textbf{+5.7} & \cint{+3.0}{+8.3} \\  %
Qwen3-VL-8B & +7.9 & +4.1\rlap{$^\dagger$} & 45.1 & 43.4 & $-$1.7 & \cint{$-$4.5}{+1.0} & +14.2 & 27.3 & 34.2 & \textbf{+6.9} & \cint{+4.0}{+9.9} \\  %
Qwen3-VL-32B & +10.5 & +6.2 & 50.7 & 50.9 & +0.2 & \cint{$-$2.4}{+3.0} & +17.8 & 30.7 & 37.8 & \textbf{+7.1} & \cint{+4.0}{+10.2} \\  %
InternVL3.5-8B & +13.2 & +17.3 & 39.2 & 40.8 & +1.6 & \cint{$-$1.9}{+5.0} & +11.2 & 35.9 & 34.9 & $-$1.0 & \cint{$-$4.3}{+2.1} \\  %
Gemma-3-12B & +5.3 & +6.0 & 37.5 & 36.3 & $-$1.2 & \cint{$-$4.2}{+1.9} & +4.5 & 33.7 & 33.7 & 0.0 & \cint{$-$3.0}{+3.0} \\  %
GLM-4.1V-9B-Th. & +2.6 & +3.6 & 56.3 & 53.6 & $-$2.6 & \cint{$-$5.6}{+0.3} & -- & 28.8 & 25.9 & $-$3.0 & \cint{$-$5.9}{+0.0} \\  %
Glyph & $-$2.6 & +2.5 & 52.8 & 48.3 & $-$4.5 & \cint{$-$8.0}{$-$1.2} & -- & -- & -- & -- & \\  %
\bottomrule
\end{tabular}
\end{table*}

\begin{table}[!ht]
\caption{\textbf{The ranking holds across families and analyses.} Spearman $\rho$ between a predictor and the representation benefit $\Delta$ on MMLongBench-Doc.}
\label{tab:app-rank}
\centering\footnotesize
\setlength{\tabcolsep}{4pt}
\setlength{\aboverulesep}{0pt}\setlength{\belowrulesep}{0pt}
\renewcommand{\arraystretch}{1.12}
\begin{tabular}{p{0.56\linewidth}rl}
\toprule
\headrow \textbf{Predictor and points} & \multicolumn{1}{c}{$\boldsymbol{\rho}$} & \textbf{Test} \\
\midrule
\multicolumn{3}{l}{\textit{Registered}} \\
15\,px gap, eight readers at 144\,DPI (Addendum 1) & $+0.86$ & exact one-sided $p = 0.005$ \\
\quad document bootstrap (130 documents) & & 95\% CI $[+0.24, +0.95]$ \\
Matched gap, 13 points over both densities (Addendum 2) & $+0.79$ & permutation $p = 0.001$ \\
\midrule
\multicolumn{3}{l}{\textit{Post hoc, same data}} \\
15\,px gap, four Qwen readers & $+0.95$ & \\
15\,px gap, the four other readers & $+1.00$ & \\
\quad both families, stratified & & exact one-sided $p = 0.003$ \\  %
Matched gap within 144\,DPI ($n = 7$) / 48\,DPI ($n = 6$) & $+0.61$ / $+0.49$ & \\
\quad both densities, stratified & & exact one-sided $p = 0.035$ \\  %
15\,px gap, recognition by second transcript (94 questions) & $+0.66$ & exact one-sided $p = 0.042$ \\
15\,px gap vs.\ words $-$ pixels over all questions & $+0.71$ & $p = 0.03$ \\
Recognition rate, 144\,DPI / 13 points & $+0.38$ / $-0.53$ & \\
\bottomrule
\end{tabular}
\end{table}

\textbf{Results by registration.} Original sign rule: correct for 1 of the 3 readers it named that were run (Glimpse was not). Addendum 1: ranking $\rho = +0.86$ (pass); sign rule 3 of 7 (fail). Addendum 2: confirmatory $\rho = +0.79$ (pass); sign rule 11 of 14. Within the 48\,DPI arm, InternVL tiles each small page into twelve 448-pixel crops, so its second look is an enlarged re-look rather than the same pixels.

\subsection{Compressed Documents: Absolute Accuracy and Cost}\label{app:abs}
\label{app:cost}
\Cref{tab:app-abs} gives the absolute accuracy and effective compression rate behind each same-budget gain in \cref{tab:main}.

\begin{table}[ht]
\caption{\textbf{Absolute accuracy behind \cref{tab:main}.} SubEM and effective compression rate (original-text tokens over the tokens in the final context), averaged over HotpotQA, MuSiQue and NQ. The frozen Qwen3-VL-8B reading the full original text scores 58.5.}
\label{tab:app-abs}
\centering\footnotesize
\setlength{\tabcolsep}{9pt}
\setlength{\aboverulesep}{0pt}\setlength{\belowrulesep}{0pt}
\renewcommand{\arraystretch}{1.12}
\begin{tabular}{lrrrrrr}
\toprule
\headrow & \multicolumn{2}{c}{\textbf{5$\times$}} & \multicolumn{2}{c}{\textbf{10$\times$}} & \multicolumn{2}{c}{\textbf{15$\times$}} \\
\cmidrule(lr){2-3}\cmidrule(lr){4-5}\cmidrule(lr){6-7}
\headrow \textbf{Method} & \textbf{Acc.} & \textbf{Rate} & \textbf{Acc.} & \textbf{Rate} & \textbf{Acc.} & \textbf{Rate} \\
\midrule
Uniform rendering & 27.7 & 5.9$\times$ & 18.3 & 11.7$\times$ & 16.5 & 16.4$\times$ \\
\midrule
\textcolor{demph}{AGAR} & \textcolor{demph}{33.3} & \textcolor{demph}{5.4$\times$} & \textcolor{demph}{20.5} & \textcolor{demph}{10.7$\times$} & \textcolor{demph}{17.8} & \textcolor{demph}{15.2$\times$} \\
\textcolor{demph}{VERA (text)} & \textcolor{demph}{48.0} & \textcolor{demph}{4.3$\times$} & \textcolor{demph}{40.7} & \textcolor{demph}{4.9$\times$} & \textcolor{demph}{30.2} & \textcolor{demph}{5.6$\times$} \\
\textcolor{demph}{LensVLM-9B$^\dagger$} & \textcolor{demph}{56.5} & \textcolor{demph}{4.1$\times$} & \textcolor{demph}{51.7} & \textcolor{demph}{5.7$\times$} & \textcolor{demph}{37.2} & \textcolor{demph}{7.9$\times$} \\
\textcolor{demph}{BM25} & \textcolor{demph}{56.5} & \textcolor{demph}{5.5$\times$} & \textcolor{demph}{52.8} & \textcolor{demph}{9.7$\times$} & \textcolor{demph}{49.7} & \textcolor{demph}{12.8$\times$} \\
\midrule
\textcolor{demph}{\ours, every page} & \textcolor{demph}{55.0} & \textcolor{demph}{5.5$\times$} & \textcolor{demph}{43.5} & \textcolor{demph}{9.9$\times$} & \textcolor{demph}{34.5} & \textcolor{demph}{13.5$\times$} \\
\textcolor{demph}{DeepSeek-OCR + BM25} & \textcolor{demph}{56.5} & \textcolor{demph}{5.5$\times$} & \textcolor{demph}{43.7} & \textcolor{demph}{9.8$\times$} & \textcolor{demph}{22.2} & \textcolor{demph}{13.2$\times$} \\
\midrule
Same reader & 57.5 & 0.9$\times$ & 50.3 & 0.9$\times$ & 39.3 & 0.7$\times$ \\
DeepSeek-OCR & 57.0 & 0.9$\times$ & 48.3 & 0.7$\times$ & 26.8 & 0.3$\times$ \\
Thinking reader & 21.8 & 0.9$\times$ & 22.2 & 2.3$\times$ & 22.7 & 3.3$\times$ \\
Zoom agent & 28.0 & 5.6$\times$ & 24.0 & 9.4$\times$ & 19.3 & 11.6$\times$ \\
DeepEyes$^\dagger$ & 27.2 & 1.6$\times$ & 19.7 & 1.6$\times$ & 16.0 & 1.6$\times$ \\
ViCrop & 38.2 & 5.2$\times$ & 24.7 & 9.0$\times$ & 19.3 & 11.7$\times$ \\
VERA (pixels) & 43.0 & 4.8$\times$ & 30.5 & 6.9$\times$ & 22.8 & 8.1$\times$ \\
Glyph$^\dagger$ & 37.8 & 4.7$\times$ & 28.0 & 8.9$\times$ & 25.0 & 11.2$\times$ \\
\midrule
\oursrow \ours (1 page) & 42.3 & 19.8$\times$ & 31.2 & 15.9$\times$ & 23.5 & 22.3$\times$ \\
\oursrow \textbf{\ours} & 54.8 & 7.2$\times$ & 39.3 & 10.6$\times$ & 28.3 & 14.0$\times$ \\
\bottomrule
\end{tabular}
\end{table}

\begin{table}[!ht]
\caption{\textbf{Nearly all the cost of \ours is reading aloud.} Tokens per question, means over the 600 test questions of each level, Qwen3-VL-8B. Pre.: prefilled; Gen.: generated. The found pages are three, three and four, enlarged to 15\,px. Longest pass: the most tokens a single forward holds, prefill and generation.}
\label{tab:app-cost}
\centering\footnotesize
\setlength{\tabcolsep}{3.5pt}
\setlength{\aboverulesep}{0pt}\setlength{\belowrulesep}{0pt}
\renewcommand{\arraystretch}{1.12}
\begin{tabular}{lrrrrrr}
\toprule
\headrow & \multicolumn{2}{c}{\textbf{5$\boldsymbol{\times}$}} & \multicolumn{2}{c}{\textbf{10$\boldsymbol{\times}$}} & \multicolumn{2}{c}{\textbf{15$\boldsymbol{\times}$}} \\
\headrow \textbf{Step} & \textbf{Pre.} & \textbf{Gen.} & \textbf{Pre.} & \textbf{Gen.} & \textbf{Pre.} & \textbf{Gen.} \\
\midrule
Uniform rendering & 1{,}675 & 4 & 897 & 5 & 729 & 5 \\  %
\midrule
Scan: all pages & 1{,}675 & -- & 897 & -- & 729 & -- \\  %
Read aloud: found pages & 955 & 1{,}347 & 991 & 1{,}799 & 988 & 2{,}077 \\  %
Answer: kept passages & 1{,}276 & 5 & 881 & 6 & 676 & 6 \\  %
\oursrow \textbf{\ours, total} & 3{,}905 & 1{,}353 & 2{,}769 & 1{,}805 & 2{,}393 & 2{,}083 \\  %
\midrule
Longest pass & \multicolumn{2}{c}{1{,}692} & \multicolumn{2}{c}{1{,}121} & \multicolumn{2}{c}{1{,}060} \\  %
\bottomrule
\end{tabular}
\end{table}

\Cref{tab:app-cost} counts the tokens of each step of \ours over all 600 test questions at each level. Nearly all the compute is reading aloud: 1{,}347, 1{,}799 and 2{,}077 generated tokens per question at $5\times$, $10\times$ and $15\times$, against about 3{,}800 for the thinking reader and 9{,}500 for transcribing every page at $10\times$. No pass is long: the longest holds 1{,}692, 1{,}121 and 1{,}060 tokens on average, a fifth, an eighth and an eighth of the original text (8{,}847 tokens). The reader answers from 1{,}232, 839 and 635 tokens of kept passages. Timed one question at a time on one RTX 6000D with vLLM, reading aloud in strips takes 1.5, 1.6 and 2.2\,s at $5\times$, $10\times$ and $15\times$ (whole pages: 6.7, 9.4 and 11.7\,s), answering 0.2\,s and the scan about 0.2\,s.  %
Counted the same way, more pixels buy less: at $5\times$, \ours answers 54.8 with 5.3K tokens in all, while uniform rendering answers 48.5 from 5.4K image tokens (paired $+6.3$ $[+2.3, +10.2]$) and 52.7 from 13.3K, more than the original text holds ($+2.2$ $[-1.7, +6.0]$).  %

\subsection{From Pixels to Words}\label{app:ablation}
With the same scan, the more of the evidence the reader receives in its own words, the more it answers (\cref{tab:app-ablation}a): pages plus found lines in pixels, then the lines read aloud, then the found pages read aloud alone. Pointing at the evidence does nothing; re-encoding it does (\cref{tab:app-ablation}c and the gold rows of a).

\begin{table*}[!ht]
\caption{\textbf{From pixels to words.} Qwen3-VL-8B on the compressed long documents of \cref{tab:main}. (a) Same-budget gain over uniform rendering, mean over the three datasets. First pass: lines scored by the scan; closer look: rows rescored after the top pages are enlarged and appended. (b) Recall of gold lines among the top six, first pass$\to$closer look, and share of gold pages among the pages read (three at $5\times$ and $10\times$, four at $15\times$). (c) Accuracy given only the gold lines, $10\times$, 50 questions per dataset. (d) Accuracy with the same selected lines, 600 questions; by confidence: the answer with the higher mean token log-probability. Gold rows: \ours. \textbf{Bold}: best in the column, excluding the rows with gold evidence or above the cost cap in (a) and the original text in (c).}
\label{tab:app-ablation}
\centering\footnotesize
\setlength{\aboverulesep}{0pt}\setlength{\belowrulesep}{0pt}
\renewcommand{\arraystretch}{1.12}
\begin{minipage}[t]{0.51\textwidth}\vspace{0pt}\centering
\textbf{(a)} What the reader answers from: same-budget gain\\[3pt]
\setlength{\tabcolsep}{5pt}
\begin{tabular}{lrrr}
\toprule
\headrow & \multicolumn{1}{c}{\textbf{5$\boldsymbol{\times}$}} & \multicolumn{1}{c}{\textbf{10$\boldsymbol{\times}$}} & \multicolumn{1}{c}{\textbf{15$\boldsymbol{\times}$}} \\
\midrule
\multicolumn{4}{l}{\textit{Pages $+$ six found lines}} \\
First pass, enlarged & 12.2 & 7.9 & 2.4 \\  %
Closer look, enlarged & 19.2 & 15.8 & 9.3 \\  %
\quad cropped to text lines & 17.2 & 15.8 & 10.1 \\  %
\quad read aloud & 21.7 & \textbf{19.9} & 7.3 \\  %
\midrule
\multicolumn{4}{l}{\textit{Pages read aloud, words only}} \\
Top page & 25.8 & 14.4 & 7.0 \\  %
\oursrow Found pages, 3 (4 at $15\times$) = \ours & \textbf{30.9} & 19.8 & \textbf{10.9} \\  %
Every page, no scan (above the cap) & 26.5 & 23.2 & 16.8 \\  %
\midrule
\multicolumn{4}{l}{\textit{Gold evidence}} \\
Gold pages read aloud & 37.6 & 26.1 & 17.7 \\  %
Gold lines, box drawn & 0.0 & 1.2 & 0.0 \\  %
Gold lines, copied in pixels & 5.4 & 1.7 & 1.3 \\  %
Gold lines, enlarged & 29.6 & 25.6 & 20.9 \\  %
\bottomrule
\end{tabular}
\end{minipage}\hfill
\begin{minipage}[t]{0.47\textwidth}\vspace{0pt}\centering
\textbf{(b)} Localization: gold lines and pages found\\[3pt]
\setlength{\tabcolsep}{6pt}
\begin{tabular}{lccc}
\toprule
\headrow & \textbf{HQA} & \textbf{MSQ} & \textbf{NQ} \\
\midrule
\multicolumn{4}{l}{\textit{Gold-line recall@6}} \\
5$\times$ & .53$\to$.70 & .28$\to$.38 & .41$\to$.54 \\  %
10$\times$ & .37$\to$.69 & .19$\to$.35 & .33$\to$.54 \\  %
15$\times$ & .15$\to$.39 & .11$\to$.27 & .23$\to$.44 \\  %
\midrule
\multicolumn{4}{l}{\textit{Gold pages among the pages read}} \\
5$\times$ & .88 & .62 & .75 \\  %
10$\times$ & .78 & .61 & .69 \\  %
15$\times$ & .44 & .39 & .53 \\  %
\bottomrule
\end{tabular}
\end{minipage}

\vspace{10pt}
\begin{minipage}[t]{0.51\textwidth}\vspace{0pt}\centering
\textbf{(c)} Gold lines: accuracy\\[3pt]
\setlength{\tabcolsep}{5pt}
\begin{tabular}{lccc}
\toprule
\headrow \textbf{Gold lines given as} & \textbf{HQA} & \textbf{MSQ} & \textbf{NQ} \\
\midrule
Enlarged token rows & 46.0 & 28.0 & 50.0 \\  %
Aligned line crop & 46.0 & 40.0 & 56.0 \\  %
\oursrow Aligned line crop, read aloud & \textbf{60.0} & \textbf{46.0} & \textbf{64.0} \\  %
\midrule
Original text & 66.0 & 52.0 & 76.0 \\  %
\bottomrule
\end{tabular}
\end{minipage}\hfill
\begin{minipage}[t]{0.47\textwidth}\vspace{0pt}\centering
\textbf{(d)} Selected lines: accuracy\\[3pt]
\setlength{\tabcolsep}{5pt}
\begin{tabular}{lccc}
\toprule
\headrow \textbf{Selected lines given as} & \textbf{5$\boldsymbol{\times}$} & \textbf{10$\boldsymbol{\times}$} & \textbf{15$\boldsymbol{\times}$} \\
\midrule
Crop only & 46.7 & 37.2 & 28.7 \\  %
\oursrow Transcript only & 52.0 & \textbf{43.7} & 29.0 \\  %
Crop and transcript & \textbf{53.8} & 42.3 & \textbf{30.8} \\  %
By confidence & 50.8 & 43.5 & 30.7 \\  %
\bottomrule
\end{tabular}
\end{minipage}
\end{table*}

\textbf{What the reader answers from.} With the same scan, \cref{tab:app-ablation}a moves from pixels to words. Scoring lines in the first pass and enlarging the top six next to the compressed pages gives $+12.2$/$+7.9$/$+2.4$ at $5\times$/$10\times$/$15\times$; looking again at the top pages enlarged, then rescoring their rows, raises this to $+19.2$/$+15.8$/$+9.3$ and the recall of gold lines among the top six by $1.3$--$2.6\times$ (\cref{tab:app-ablation}b). Reading those lines aloud next to the pages gives $+21.7$/$+19.9$/$+7.3$. Reading the found pages aloud and answering from the words alone, which needs no second look, gives $+30.9$/$+19.8$/$+10.9$; reading only the top page still gives $+25.8$/$+14.4$/$+7.0$. Above the cost cap, for reference: reading every page aloud, without the scan, and ranking its passages by BM25 gives $+26.5$/$+23.2$/$+16.8$ for 9{,}389, 9{,}478 and 11{,}921 generated tokens per question; reading the found pages takes a seventh to a fifth of those words, and is ahead at $5\times$ (paired $+4.5$ $[+1.2, +7.6]$) and behind at $10\times$ and $15\times$ ($-3.4$ $[-6.8, -0.3]$, $-5.9$ $[-9.2, -2.6]$). DeepSeek-OCR on every compressed page with BM25 gives $+27.8$/$+23.2$/$+4.4$; \ours is ahead at $5\times$ and $15\times$ (paired $+3.1$ $[-0.2, +6.6]$, $+6.5$ $[+2.9, +10.2]$) and behind at $10\times$ ($-3.4$ $[-7.2, +0.3]$). At $15\times$ its transcripts of the unenlarged pages degenerate into repetition, about 26K tokens per question. %
 Reading the gold pages aloud bounds what locating can give ($+37.6$/$+26.1$/$+17.7$). A linear readout over all heads beats selecting the eight best heads (short setting, HotpotQA, three compression levels: $+9.3$/$+13.3$/$+7.8$ vs.\ $+6.2$/$+8.0$/$+4.3$).

\textbf{Localization and recognition.} The pages the scan reads cover 75\%, 69\% and 45\% of the gold pages at $5\times$, $10\times$ and $15\times$, against 16\%, 21\% and 20\% for as many random pages, and the top three pages of whole MMLongBench-Doc documents at 48\,DPI cover 77\% (top one, two and five: 54\%, 71\%, 83\%). At $10\times$ the reader recognizes 47\% of the evidence at its native height. Its transcripts of the found pages have a character similarity to the original text of 0.90--0.98, 0.61--0.79 and 0.36--0.56 at $5\times$, $10\times$ and $15\times$. %

\textbf{Second look.} On gold lines at $10\times$ (50 questions per dataset), enlarging the gold token rows gives 46.0/28.0/50.0 (HotpotQA/MuSiQue/NQ), cropping to aligned text lines 46.0/40.0/56.0, transcribing those lines 60.0/46.0/64.0, and the original text 66.0/52.0/76.0. Pointing is not enough: drawing a box around the gold lines gives $+1.2$/$+0.0$ at $10\times$/$15\times$ and copying them in pixels $+1.7$/$+1.3$, while enlarging them gives $+25.6$/$+20.9$.

\textbf{The same passages as pixels.} We crop the passages \ours keeps from the compressed pages (each passage's rows located by the same proportional alignment that cut the transcript into passages), stack them, enlarge them to 15\,px by bicubic interpolation and hand the reader only this image. At $5\times$, $10\times$ and $15\times$ it answers 44.0, 31.2 and 23.8 against 54.8, 39.3 and 28.7 from the words of the same passages in the same run (paired $+10.8$ $[+7.5, +14.3]$, $+8.2$ $[+4.7, +11.7]$, $+4.8$ $[+1.8, +8.0]$), from 53\%, 42\% and 40\% of the tokens. Cut, in \ours's ranking, to the pixels' token count, the words answer as well from 85\%, 75\% and 65\% of those tokens (45.3, 29.8 and 23.3; paired with the pixels $+1.3$ $[-2.3, +4.8]$, $-1.3$ $[-4.7, +2.0]$, $-0.5$ $[-3.5, +2.5]$). Per token, pixels hold more of the passages and words make more of what they hold; at equal tokens the two balance. %

\textbf{The whole budget as pixels.} Given the budget of \ours instead, the pixels hold the passages \ours keeps and then blocks of the other pages, of the same number of lines and taken in the order of their scan scores, stacked in document order and enlarged until they reach the budget; no further page is transcribed. They answer 46.3, 34.3 and 25.2, using 98\%, 97\% and 96\% of the budget against the words' 73\%, 85\% and 84\% (paired with the words $+8.5$ $[+4.5, +12.5]$, $+5.0$ $[+1.3, +9.0]$, $+3.5$ $[+0.2, +7.0]$). From 647 to 1{,}655 visual tokens at $5\times$, the pixels gain 2.3 points; from 550 to 1{,}232 text tokens, the words gain 9.5. %

\textbf{Words, pixels, or both.} With the same selected lines, giving only the transcript, the transcript with the enlarged crop, or choosing by the reader's confidence differ by at most 3 points at each level, while the enlarged crop alone trails them by 4--7 points at $5\times$ and $10\times$ and matches them at $15\times$, where the transcripts are unreliable.

\subsection{Transfer and the Form of the Evidence}\label{app:forms}
\begin{table}[!ht]
\caption{\textbf{Words are at least as good on rendered text; pages win on native scans} (accuracy). W: the frozen Qwen3-VL-8B's words (the passages \ours keeps; for MMLongBench-Doc, its transcript of the two scanned pages); P: the same evidence as enlarged pixels (the kept passages cropped from the pages enlarged to 15\,px; the two scanned pages at 144\,DPI). Handing over the three found pages whole at 15\,px instead gives 58.3 (GPT-6.1) and 78.0 (Gemini~4) with about twice the tokens of W. Frontier rows: 60 questions. \textbf{Bold}: the form \ours hands over.}
\label{tab:app-forms}
\centering\footnotesize
\setlength{\tabcolsep}{3pt}
\setlength{\aboverulesep}{0pt}\setlength{\belowrulesep}{0pt}
\renewcommand{\arraystretch}{1.12}
\begin{tabular}{llrrl}
\toprule
\headrow \textbf{Setting} & \textbf{Reader} & \textbf{W} & \textbf{P} & \textbf{Why} \\
\midrule
MMLongBench-Doc & Qwen3-VL-8B & 42.7 & \textbf{45.6} & native 144\,DPI, about 24\,px \\  %
Long-doc QA, $10\times$ & GPT-6.1 & \textbf{55.0} & 51.7 & rendered text \\  %
Long-doc QA, $10\times$ & Gemini 4 & \textbf{81.7} & 76.7 & rendered text \\  %
\bottomrule
\end{tabular}
\end{table}

On RULER the gain holds at every length: Glyph's 73.8, 72.2, 74.8, 73.6, 73.3 and 71.4 at 4K, 8K, 16K, 32K, 64K and 126K tokens become 79.2, 78.8, 82.5, 80.5, 80.4 and 77.3. %
\ours reads aloud everywhere except where the found pages, at their native resolution, already reach the character height at which the reader's answers from the image catch up (\cref{app:impl}). On the scanned pages of MMLongBench-Doc at 144\,DPI the pages win, and no reader of \cref{sec:predict} gains significantly from words there; on rendered text, words are at least as good for both frontier readers, whether or not they have a gap. The gain from words over the compressed pages alone is $+6.7$ (paired 95\% CI $[+3.5, +9.8]$) for Gemini~4 on 600 questions and $+1.7$ ($[-8.5, +11.9]$) for GPT-6.1 on the 59 questions it answered under both conditions ($+2.5$ over all 60 in \cref{tab:transfer}). With its own zoom tool, Gemini~4 reaches 79.3 at 6{,}680 billed tokens; given our reader's words, 81.7 at 2{,}015, as from the original text (81.7 at 8{,}938). Qwen3-VL-4B, with its own scan and transcripts, gains $+30.4$, $+18.7$ and $+5.8$ at $5\times$, $10\times$ and $15\times$ (HotpotQA/MuSiQue/NQ: $+34.6$/$+20.7$/$+35.9$, $+19.1$/$+10.5$/$+26.5$, $+5.7$/$+1.0$/$+10.6$). %

\section{Supplementary Information}\label{app:si}

\subsection{Prompts and Decoding}\label{app:prompts}
\textbf{Prompts.} Transcription, for every reader and every page: ``Transcribe all the text in this image exactly, line by line. Output only the transcribed text.'' In \cref{sec:gap}, the evidence (text, image or transcript) follows ``Here is an excerpt of the document:'' and is followed by ``Question: \{question\}'' and ``Answer with a short phrase only.'' \ours answers from ``Document excerpts:'' followed by the kept passages and the same two lines; uniform rendering gives the pages, each after ``Image $k$:'', in their place. Decoding is greedy throughout, except for the thinking reader of \cref{tab:main}, which uses its recommended sampling.

\subsection{Release}
\textbf{Release.} Code, prompts, the registrations of \cref{app:prereg} and per-question outputs will be released with the paper.

\end{document}